\documentclass{amsart}

\usepackage{amsmath, amsfonts,amssymb,amsthm,graphicx} 
\usepackage{natbib} 

\newtheorem{proposition}{Proposition}

\usepackage[norelsize]{algorithm2e} 

\newcommand{\R}{\mathbb{R}}

\begin{document}


\title{Bayesian learning of noisy Markov decision processes}
\author{Sumeetpal S. Singh}
\address{University of Cambridge}
\email{sss40@cam.ac.uk}
\author{Nicolas Chopin}
\address{CREST---ENSAE and HEC Paris}
\email{nicolas.chopin@ensae.fr}
\author{Nick Whiteley}
\address{ University of Bristol}
\email{Nick.Whiteley@bristol.ac.uk}

\begin{abstract}
We consider the inverse reinforcement learning problem, that is, the problem of learning from, and then predicting or mimicking a controller based on state/action data. 
We propose a statistical model for such data, derived from the
structure of a Markov decision process. Adopting a Bayesian approach to
inference, we show how latent variables of the model can be estimated, and how predictions about actions can be made, in a unified
framework. A new Markov chain Monte Carlo (MCMC) sampler is devised for simulation from the posterior distribution. This step includes a parameter expansion step, which is shown to be essential for good convergence properties of the MCMC sampler.  As an illustration, the method is applied to learning a human controller.

\end{abstract}
%
%
%
%
%

\maketitle

\section{Introduction}
\subsection{Motivation}

The problem of fitting a statistical model to observed actions has
received significant attention in a variety of disciplines. These
include Optimal Control \citep{Rus88}, Economics
\citep{GoM80,Wol84,Rus87,HoM93,GeK00,GKR94,AgM02,IJC09},  and
Machine Learning \citep{NgR00,AbN04}. Across these cases there is
some variety in the estimation aims and the assumed mechanisms
which generate observed actions. We focus on the case in which it
is assumed that the observations arise from an underlying Markov
Decision Process (a full model specification is given in the next
section). In this case, given the current state $X$ of the system,
the controller chooses an action $A$ and receives an instantaneous
reward specified by the function $r$,
\begin{equation*}
(X,A)\rightarrow r(X,A)\in \mathbb{R}
\end{equation*}%
where $(X,A)$ is the state-action pair. The state then evolves according to
a Markov kernel $p(\cdot |X,A)$, the controller chooses the next action,
receives a reward and
so on.
The controller chooses its actions to maximize the average reward it will accrue
over an
infinite horizon. However, the controller may take sub-optimal
decisions from time to time. This is captured by a \textquotedblleft
noisy\textquotedblright\ Markov Decision Process (MDP) model.

A generic approach to automating a task is to model it as a control problem, see Chapter 8 of \citet{BeT96}, and also \citet{bertsekas:bookvol1},  
\citet{bertsekas:bookvol2}, 
which involves specifying a reward function and other
elements of the model, and then solving for the optimal
controller. Putting aside the difficulties associated with the
last step, specifying the reward function is non-trivial and often
achieved in practice by a heuristic process of trial and error,
i.e. by observing the system under the computed optimal controller
and then adjusting the reward function to avoid observed undesired
behavior. After this adjustment, the optimal controller is
re-computed and this process is repeated until satisfaction.

An alternative, often simpler approach is to obtain a sample path which is
characteristic of desired behavior, e.g. by a human controller,
and then estimate the control policy generating this sample path:
this is known as the inverse reinforcement learning problem
\citep{NgR00,AbN04}. Learning to mimic a controller has many
potential applications in applied fields such as robotics and
artificial intelligence \citep{CAN09}; biology, e.g. the study of
animal learning \citep{Wat87,ScZ97}, economics \citep{Rus87}, and
other fields.

Our aim is to develop a purely statistical, and computationally
tractable, solution to this problem of mimicking behaviour based on a statistical model
for the controller's actions, and the Bayesian approach. This is
advantageous because it gives us a unified and principled
framework for the following tasks: (a) to properly model
uncertainty (i.e.  the human controller may make mistakes); (b) to
estimate jointly the control policy and the model parameters, (c)
to predict future actions.

In a parametric approach to estimation, the reward function and
other elements of the model are assumed to be specific functional
forms of a parameter vector $\theta$
\citep{GoM80,Wol84,Rus87,Rus88,AgM02,HoM93,IJC09}. The best
parametric estimate may then be computed, for example, by
maximizing the likelihood of the observed data with respect to
$\theta$. From a computational perspective, we shall see that this
approach is cumbersome for a noisy MDP model because the
likelihood of an observed action given observed states is an intractable
integral. We avoid this difficulty by targeting the
control policy directly; as we shall see, a specific quantity
called the optimal value function. This gives an additional
justification to the Bayesian approach in this context, as the
data augmentation principle \citep{TaW87} makes it possible to
estimate the model without computing the difficult integral
mentioned.

\subsection{Contributions}

The contributions of this work are as follows. We adopt a Bayesian
approach to modelling state/action data derived from the structure of an MDP.
Our
approach is inspired by the pioneering work of
\citet{AlC93} and \citet{McR94} in the context of statistical inference
in discrete choice models, where the computations are performed
using a Gibbs sampler applied to an enlarged (or augmented) model.
In subsequent work, \citet{Nob98} and \citet{ImV05} enhanced the
computational efficiency of the Gibbs sampling technique while
\citet{MPR00} and \citet{ImV05} devised new priors for the identified
parameters of the model.

We devise a new Gibbs sampling algorithm for inferring the 
optimal value function. The proposed algorithm is a Parameter Expanded
Data Augmentation (PX-DA) algorithm \citep{LiW99,MeV99}. PX-DA
improves upon the efficiency of standard DA by reducing the
correlation between the samples. This is achieved by inserting an
additional simulation step into the algorithm which involves
moving in the augmented data space.
This extra simulation step is computationally inexpensive and
leads to improved performance over standard DA algorithms. In
fact, we give examples where the DA algorithm does not converge after a large
number of iterations,
whereas PX-DA does. The PX-DA algorithm we propose involves movement of the 
augmented data in the extra simulation
step with a combination of translation and scaling. We also implement
an efficient Metropolis-Hastings kernel with independent proposals
when sampling the augmented data.

As an illustrative example of learning a human controller we apply
our framework to the game of Tetris. Automating Tetris is a
challenging benchmark problem in the control literature, see for
example \citet{BeT96}, and treating it is difficult because the
control model has a very large state space. Moreover, data from a
human player is noisy, as we are prone to making errors. We show
the proposed method can quite accurately mimic a given human player by
performing posterior prediction from a limited amount of observed
data from that player. By contrast, existing approaches from the control literature
focus solely on, having specified the reward function,
solving the associated difficult optimization problem using
reinforcement learning or other dynamic programming algorithms \citep{bertsekas:bookvol1, bertsekas:bookvol2, TsV94,BeT96}. 
We also demonstrate the effect of the amount of data available from a player on
posterior distributions which, under the proposed model, characterise their
action preferences.

\subsection{Plan, notation}

The organization of this paper is as follows. Section
\ref{sec:problemState} defines the problem in detail and states the
inference objectives. Section \ref{sec:pxda} describes the PX-DA
method generally and then the specific implementation for the model we consider.
Section \ref{sec:implementation} presents the
PX-DA sampler in detail for the assumed priors and discusses some
practical issues and extensions. Numerical results highlighting
various properties of the proposed PX-DA algorithm are presented in
Section \ref{sec:experiments}, as well as implementation details and
results for Tetris. A proof of the correctness of the proposed PX-DA
algorithm is presented in the Appendix along with implementation
details of the MCMC algorithm.

This section is concluded with a description of the notation used.
Capital letters are used for random variables and lower case for
their realizations. We use the colon short-hand
for sequence of random variables, e.g. $X_{0:k}=(X_0,\ldots,X_k)$.
The letters $f$ and $p$ are reserved for the
probability densities or probability mass functions of random
variables. For two jointly distributed random variables $(X,Y)$,
$f_{X|Y}$, $f_{X,Y}$ and $f_{X}$ denote, respectively, the
conditional probability density, the joint density and the
marginal density. When the subscript is omitted, the arguments of
$f$ or $p$ will indicate precisely the random variables to which
the density corresponds. For example, $p(x|y)$ is $p_{X|Y}(x|y)$.
The value at $x$ of the
multivariate normal probability density with mean $\mu$ and
covariance $\Sigma$ is denoted $\mathcal{N}(x;\mu ,\Sigma )$. 
For a vector $v$, the $i$-th component is
denoted $v(i)$. All vectors are column vectors and the transpose
of $v$ is indicated by $v^{\text{T}}$. The $m$-dimensional vector
comprised of ones
(respectively zeros) only is denoted by $\mathbf{1}_{m}$ (respectively $%
\mathbf{0}_{m}$). The subscript $m$ is omitted when the dimension is obvious
from context.\ $I_{m}$\ will denote the $m$ by $m$ identity matrix.
Similarly, $[\Sigma ]_{i,j}$ will denote the $(i,j)$-th element of the
matrix $\Sigma $. $\mathbb{I}_{A}$ is the indicator function of the set $A$,
i.e. $\mathbb{I}_{A}(x)=1$ if $x\in A$\ and 0 otherwise. $\mathbb{R}$
denotes the real line, $\mathbb{R}_{+}$ its strictly positive part and $%
\mathbb{E}$ is the mathematical expectation operator. The cardinality of a
finite set $A$ is denoted by $\left\vert A\right\vert $. The Dirac measure
concentrated at a point $x$ is denoted by $\delta_x$.

\section{Problem Statement}\label{sec:problemState}

\subsection{Markov decision processes}

An MDP is comprised of a controlled Markov chain, a control
process, a reward function and an optimality criterion. Each of
these are defined in turn below; see \citet{bertsekas:bookvol1, bertsekas:bookvol2} for additional
background and details of other MDP optimality criteria.

The state process, denoted $\{X_{t}\}_{t\geq 1}$, is a $\mathcal{X}$-valued\
controlled discrete time (so $t$
is always an integer) Markov chain where $\mathcal{X}$\ is the finite set $%
\{1,2,\ldots,N\}$. Let $\{A_{t}\}_{t\geq 1}$ be the $\mathcal{A}$-valued
control (or action) process where $\mathcal{A}=\{1,2,\ldots,M\}$ is the set
of all possible controls. Given the entire realization of the state and
actions up to time $t\geq 1$, the evolution of the state to time $t+1$\ is
determined by the selected action and state at time $t$ only, i.e.
\begin{equation}
\left. X_{t+1}\right\vert \left( X_{1:t}=x_{1:t},\,A_{1:t}=a_{1:t}\right)
\sim p(\cdot|x_{t},a_{t}),  \label{eq:stateModel}
\end{equation}
where for each state-action pair $(x,a)$, $p(\cdot|x,a)$ is a probability
distribution on $\mathcal{X}$. The evolution of the action process is
determined by a policy $\mu$\ which is a mapping from the set of states to
the set of actions. Particularly, for $k\geq 1$
\begin{equation*}
\left. A_{t}\right\vert \left( X_{1:t}=x_{1:t},\,A_{1:t-1}=a_{0:t-1}\right)
\sim\delta_{\mu(x_{t})}(\cdot).
\end{equation*}
Let $r$ be a real valued function on $\mathcal{X}$ which is called the
reward function. The reward at time $t$ for being in state $X_{t}$ is $%
r(X_{t})$. We consider the following standard optimality criterion: a discounted
sum of accumulated rewards over an infinite horizon,
\begin{equation}
C_{\mu }(x_1)=\mathbb{E}_{\mu }\left[ \left. \sum\limits_{t=1}^{\infty
}\beta ^{t}r(X_{t})\right\vert X_1=x_1 \right]   \label{eq:valueFnForMu}
\end{equation}%
where $\beta \in (0,1)$\ is the discount factor ensuring the
expectation is well defined. (If there exists a zero reward state
which is absorbing, and all policies lead to this state with
probability one for all initial states $x_1$ then the
expectation is well defined, provided $\mathcal{X}$\ is a finite set, without the discount $\beta $.) The
subscript on the expectation operator denotes the policy
controlling the evolution of $\{X_{t}\}_{t\geq 1}$. 
A policy $\mu ^{\ast }$ is said to be optimal if 
$C_{\mu ^{\ast }}(x_1)\geq C_{\mu }(x_1)$ for all $(\mu,x_1).$

It is well known that $\mu ^{\ast
}$ is characterized by the real valued function on $\mathcal{X}$, denoted 
$V$, which satisfies the following fixed point equation,
\begin{equation}
V(x)=\max_{a\in \mathcal{A}}\left\{ r(x)+\beta
\sum\limits_{x^{\prime }\in \mathcal{X}}p(x^{\prime }|x,a)V(x^{\prime })\right\}
.  \label{eq:optValueFn}
\end{equation}%
In the literature on MDPs \citep{BeT96, bertsekas:bookvol1, bertsekas:bookvol2},
$V$\ is referred
to as the (optimal) value function. Since $V$ is a $\mathcal{X}\rightarrow \R$
function, 
it will be treated as a vector in $\R^N$ from now on. 
Given $V$,\ the optimal policy $\mu ^{\ast }$\ is, for all $x\in \mathcal{X}$, 
\begin{equation}
\mu ^{\ast }(x)=\arg \max_{a\in \mathcal{A}}\left\{ 
\sum\limits_{x^{\prime }\in \mathcal{X}}p(x^{\prime }|x,a)V(x^{\prime })
\right\} =
\arg \max_{a\in \mathcal{A}}\left\{ 
\left(R_t V\right) (a)
\right\}
\label{eq:optPol}
\end{equation}
where $R_t$ is a short-hand for $R_{x_t}$, and, for each $x\in\mathcal{X}$,
$R_{x}$ is the $M\times N$ transition probability matrix
with elements
\begin{equation}
\lbrack R_{x}]_{i,j}=p(j|x,i),\quad 1\leq i\leq M,\quad 1\leq j \leq
N,\label{eq:matrixR}
\end{equation}
with $p(j|x,i)$  defined as in \eqref{eq:stateModel}. 
Recall that in our notations $(R_t V)(a)$ is the $a$-th component of 
vector $R_t V$. 

\subsection{A statistical model for imperfect policy execution}

We consider the following statistical model for the action
component $a_{t}$ of each observed state-action pair
$(x_{t},a_{t})$ built around the MDP framework. It is assumed that
\begin{equation}
A_{t}=\arg \max_{a\in \mathcal{A}}\left\{ \epsilon _{t}(a)+
\left(R_t V\right)(a)
\right\}  \label{eq:obsModel}
\end{equation}%
where $R_t$ has been defined at the end of the previous section,
and the $\epsilon _{t}$'s, $t\geq 1$, are independent and identically
distributed $M$-dimensional Gaussian variates,
\begin{equation*}
\epsilon _{t}^{\text{T}}=\left( \epsilon _{t}(1),\ldots ,\epsilon
_{t}(M)\right) \sim \mathcal{N}(\mathbf{0}_{M},\Sigma ).
\end{equation*}%
(The choice of the Gaussian distribution is for computational and inferential
convenience.)  
The inclusion of this noise process renders the model more
versatile. It may be interpreted in two different ways. First, if
there are several actions that are near optimal, in the sense
quantified by the numerical value of the expression in the right
hand side of (\ref{eq:optPol}), then the controller could have
selected one of the near optimal actions in error. Thus while the
policy is optimal, the execution of the policy is subject to
disturbance. Second, it can be shown that (\ref{eq:obsModel})
characterizes the optimal policy of an MDP with a mixed
discrete-continuous state process, $(x,\epsilon )\in
\mathcal{X}\times \mathbb{R}^{M}$, and reward function
$r:\mathcal{X}\times \mathbb{R}^{M}\times \mathcal{A}\rightarrow
\mathbb{R}$ given by $r(x,\epsilon ,a)=r(x)+\epsilon (a)$.
Given the state $(x_{t},\epsilon _{t})$\ at time $t$, and action
$a_{t}$,
the discrete component of the next state, $X_{t+1}$, is drawn from
\eqref{eq:stateModel}
 while the continuous component $\epsilon _{t+1}$\
is drawn from $\mathcal{N}(\mathbf{0}_{M},\Sigma )$. It follows
from this separation in the evolution of the state components that
there exists a vector $V\in $ $\mathbb{R}^{\left\vert
\mathcal{X}\right\vert }$ such the optimal policy for this MDP is
given by \eqref{eq:obsModel} \citep[Theorems 3.1, 3.3]{Rus88}. In
this model, the statistician only observes the discrete component
of the state process and the action taken at each time, while
$\epsilon _{t}$ is the unobserved random component of the reward
known only to the decision maker.

With respect to the interpretation of the model, note that,
if $\beta$ and $V$ are fixed, then the reward function 
is entirely determined by  \eqref{eq:optValueFn}. Thus, when inferring 
$V$ from the model defined by \eqref{eq:stateModel} and \eqref{eq:obsModel}, 
(and fixing $\beta$), 
one is also implicitly inferring the optimality criterion (or equivalently
the reward function) that governed the controller's behaviour, 
and that criterion is presumably unknown to the observer, prior 
to collecting data. In practical terms, this also means that 
it remains reasonable to apply this model even when the controller's
actions seem inefficient or even erratic to the observer, as 
the observer and the controller may simply have very different 
policy criteria. 

\subsection{Inference objectives}
The data $d$ consist of a sequence of state-action pairs,
$d=\left\{ d_{t}\right\} _{t=1}^{T}=\left\{ (x_{t},a_{t})\right\}
_{t=1}^{T}$ observed for $T$ epochs and the aim is to infer
$V$. It is assumed that the law of the controlled process,
which is specified by the collection of transition matrices
$\left\{ P_{a}\right\} _{a\in \mathcal{A}}$ is known, but the
reward function $r$ is unknown. This implies (\ref{eq:optValueFn})
cannot be used to solve for $V$. The approach below can be
generalized to the case when $\left\{ P_{a}\right\} _{a\in
\mathcal{A}}$ is unknown. However, assuming $\left\{ P_{a}\right\}
_{a\in \mathcal{A}}$\ is known is reasonable in a number of
applications, in particular the human controller example studied
in Section \ref{sec:experiments}.

In Bayesian setting, a prior for $V$ is chosen and inference will be
based on samples from the posterior $p_{V|D}(v|d)$, henceforth denoted as $%
p(v|d).$ (The specification of the prior over the optimal value function is
postponed to Section \ref{sec:implementation}.) These samples may
then also be used via (\ref{eq:optPol}) to estimate the optimal
policy $\mu ^{\ast }$ and thus predict the behavior of the system.
In the context of the human controller example, $d$\ consists of
the observed actions of a person.

The likelihood of the observed data is
\begin{equation*}
p(d\left\vert v,\Sigma \right. )
=\prod\limits_{t=1}^{T}p(x_{t}|x_{t-1},a_{t-1})p(a_{t}\left\vert
v,\Sigma ,x_{t}\right. ) \propto
\prod\limits_{t=1}^{T}p(a_{t}\left\vert v,\Sigma ,x_{t}\right)
\end{equation*}%
where, abusing notation, $p(x_1|x_0,a_0)$\ denotes the prior
distribution for $X_1$. The terms 
$p(x_{t}|x_{t-1},a_{t-1})$ may be omitted as they have no bearing on the desired
posterior.

The likelihood\ $p(a_{t}\left\vert v,\Sigma ,x_{t}\right. )$, or conditional
choice probability, is the intractable integral
\begin{equation}
p(A_{t}=a_{t}\left\vert v,\Sigma ,x_{t}\right. )=\int_{\{u\in
\mathbb{R}^{M}:\, u(a_{t})\geq u(j)\mbox{ for all }j\neq a_{t}\}}\mathcal{N}%
(u;R_{t}v,\Sigma )du.  \label{eq:ccp}
\end{equation}
Henceforth $p(A_{t}=a_{t}\left\vert v,\Sigma ,x_{t}\right. )$ will be
abbreviated to $p(a_{t}\left\vert v,\Sigma \right. )$. The likelihood is
invariant to both translations of the vector $v$ and multiplications of it
by positive scalars,
\begin{equation}
p(d\left\vert v,\Sigma \right. )=p(d\left\vert \sqrt{z_{1}}(v+z_{2}\mathbf{1}%
),z_{1}\Sigma \right. ),\quad \forall (z_{1},z_{2})\in \mathbb{R}_{+}\times
\mathbb{R}.  \label{eq:invarLikelihood}
\end{equation}%
The design of the PX-DA algorithm presented in the following Section is
based on this property.

The assumed model for the noise corrupting the action selection
process results in a target distribution similar to the
multinomial probit (MNP) problem
\citep{AlC93,GKR94,McR94,MPR00,ImV05} and the stated invariance of
the likelihood to scaling ($z_{1}$) is well documented in this
literature. Specifically, in \eqref{eq:obsModel}, the 
observed actions $A_t$ correspond to the observed chosen outcomes (among 
$M$ alternatives), the random terms $\epsilon _{t}(a)$ may be
interpreted as the non-observed part of the utility function, 
and the observed part of the utility function $(R_t V)(a)$ 
may be interpreted as a linear combination of covariates, 
where the covariates are the probabilities $p(x_{t+1}=j|x_t,a_t)$,
$j=1,\ldots,N$,
and the unknown regression coefficients are 
the components of $V$. 

In the context of MNP models, the main existing
approach to ensure the posterior is well defined for improper
priors is to constrain enough parameters of the model to ensure
identifiability of the remaining ones and then introduce priors
for them. For example, by setting the last component of $V$ to
zero and then introducing a suitable prior for the remaining $N-1$
non-zero components. We
shall use a different approach as detailed in Section \ref%
{sec:implementation}.


\section{The PX-DA Method}\label{sec:pxda}

Let $f_{X}:\mathbb{R}^{p}\rightarrow\mathbb{R}$ be a target probability
density from which samples are sought. In many applications, it is not
possible to simulate from $f_{X}$ directly. However, it is often possible to
introduce a random vector $Y\in\mathbb{R}^{q}$ which is jointly distributed
with $X$ such that sampling from the conditional densities $f_{X|Y}$ and $%
f_{Y|X}$\ is straightforward. This is the principle of \emph{data
augmentation} (DA) \citep{TaW87}. Simulating from these densities
sequentially as follows,
\begin{equation}
Y_{n+1}|X_{n}=x_{n}\sim f_{Y|X}(\cdot|x_{n}),\text{\quad}%
X_{n+1}|Y_{n+1}=y_{n+1}\sim f_{X|Y}(\cdot|y_{n+1}),\text{\quad}n\geq0,
\label{eq:DA}
\end{equation}
results in a Markov chain $\{X_{n}\}_{n\geq0}$ with the correct asymptotic
distribution for any initial state $x_{0}$ (under weak regularity
assumptions) \citep{Hobert:Chapter}:
\begin{equation}
\lim_{n\rightarrow\infty}\mathbb{P}(X_{n}\in A)=\int_{A}f_{X}(x)dx.
\end{equation}
As noted by \citet{LiW99}, \citet{MeV99} in some situations it is
possible to improve the efficiency of this sampler by introducing
auxiliary variables. This technique was termed \emph{parameter
expanded (PX)} DA by \citet{LiW99}.

Let $\Lambda\subseteq\mathbb{R}^{d}$ and let $\{\varphi_{\lambda}\}_{\lambda%
\in\Lambda}$ be a class of one-to-one differentiable\ functions$\ $mapping $%
\mathbb{R}^{q}$\ to itself. Let
\begin{equation}
J_{\lambda}(y^{\prime})=\left\vert \det\left( \left[ \left. \frac {%
\partial\varphi_{\lambda,i}(y)}{\partial y(j)}\right\vert _{y=y^{\prime}}%
\right] _{i,j}\right) \right\vert  \label{eq:jacobian}
\end{equation}
where $\varphi_{\lambda,i}(y)$\ is $i$-th component function of $\varphi
_{\lambda}(y).$ $J_{\lambda}$ is the Jacobian determinant of the mapping $%
\varphi_{\lambda}:\mathbb{R}^{q}\rightarrow\mathbb{R}^{q}$. Let $Z$ be a
random vector in $\Lambda\subseteq\mathbb{R}^{d}$ with probability density $%
f_{Z}$. The aim is to reduce the auto-correlation between $X_{n}$ and $%
X_{n+1}$ generated by the Gibbs sampler and PX-DA achieves this by
inserting an extra simulation step as follows.


\noindent\hrulefill

\begin{algorithm}
\textbf{\label{alg:genPXDA} A generic PX-DA sampler}
\end{algorithm}
Given $X_{n}=x_{n}$ at iteration $n+1$, perform the following steps to sample $X_{n+1}$:

\textbf{Step 1.} Sample $Y_{n+1}$ from $f_{Y|X}(\cdot|x_{n})$ and call the
sampled value $y_{n+1}$. (If exact sampling from $f_{Y|X}$\ is not possible,
sample $Y_{n+1}$ from a Markov kernel that leaves $f_{Y|X}(\cdot|x_{n})$\
invariant.)

\textbf{Step 2a.} Sample $Z_{n+1}^{(1)}$ from $f_{Z}(\cdot)$, call the
sample $z_{n+1}^{(1)}$ and let $\widetilde{y}_{n+1}=%
\varphi_{z_{n+1}^{(1)}}^{-1}(y_{n+1})$.

\textbf{Step 2b.} Sample another $\Lambda$-valued random variable, $%
Z_{n+1}^{(2)}$, from the density which is defined (upto a proportionality
constant) by%
\begin{equation}
f_{Y}(\varphi_{z}(\widetilde{y}_{n+1}))J_{z}(\widetilde{y}_{n+1})f_{Z}(z).
\label{eq:pxda_step 2b}
\end{equation}
Call the result $z_{n+1}^{(2)}$ and set $y_{n+1}^{\prime}=\varphi
_{z_{n+1}^{(2)}}(\widetilde{y}_{n+1})$.

\textbf{Step 3.} Sample $X_{n+1}$ from $f_{X|Y}(\cdot|y_{n+1}^{\prime})$

\noindent\hrulefill

The difference between the standard DA algorithm in (\ref{eq:DA})
and PX-DA is step 2. Per iteration, PX-DA has a slightly greater
computational cost due to the need to sample the variables
$(Z_{n}^{(1)},Z_{n}^{(2)})$. However, in cases of practical
interest, these variables are typically of a much lower dimension
than $X$ or $Y$ and the increase in computational cost is often
negligible. The benefit though, in terms of the mixing rate of the
sampler, has been observed to be quite substantial in some situations
\citep{LiW99}.
Direct
simulation from the probability density on $\Lambda$\ given by (\ref%
{eq:pxda_step 2b}) is possible for the specific family of mappings $%
\{\varphi_{\lambda}\}_{\lambda\in\Lambda}$ we consider in the sequel. When a
direct draw from (\ref{eq:pxda_step 2b}) is possible the resulting PX-DA
algorithm is termed \textit{exact}.

Step 2 transforms the simulated random variable in step 1 from $y_{n+1}$ to $%
y_{n+1}^{\prime }$ via the intermediate value $\widetilde{y}_{n+1}$.
Essentially step 2 is implementing a Markov transition from $\mathbb{R}^{q}$%
\ to $\mathbb{R}^{q}$ using the kernel
\begin{equation*}
Q(y_{n+1},B)=\mathbb{E}\left\{ \mathbb{I}_{B}(Y_{n+1}^{\prime
})|Y_{n+1}=y_{n+1}\right\} =\mathbb{E}\left\{ \mathbb{I}_{B}(\varphi
_{Z_{n+1}^{(2)}}\circ \varphi
_{Z_{n+1}^{(1)}}^{-1}(Y_{n+1}))|Y_{n+1}=y_{n+1}\right\}
\end{equation*}%
It can be shown that $Q$ is reversible with respect to
the marginal distribution $f_{Y}$ of $Y$,
and thus$\ f_{Y}\ $is also invariant for $Q(y_{n},B)$
\citep[Theorem 1]{LiW99}. This in
turn implies that the invariant probability density of the Markov chain $%
\{X_{n}\}_{n\geq 0}$\ generated by the PX-DA algorithm is indeed $f_{X}$; if
$X_{n}\sim f_{X}$ then the law of $Y_{n+1}\ $is $f_{Y}$\ and, since $f_{Y}$\
is invariant for $Q$, the law of $Y_{n+1}^{\prime }\ $is also $f_{Y}$.


As was noted by \citet{LiW99}, \citet{MeV99}, it is possible to
reduce the auto-correlation between the successive $X_{n}$ samples
generated by the PX-DA algorithm by making the prior $f_{Z}$\ more
diffuse. In fact, with a trivial modification, the PX-DA algorithm
can still remain valid as an MCMC scheme when the prior is improper. The random
draw in
step 2a is then no longer well defined, but as we shall now see in the context
of a specfic transformation, the
correct procedure in this case is to omit this draw and set $\widetilde{y}%
_{n+1}$ to $y_{n+1}$ from step 1. All other steps remain unchanged. 

Let $\Lambda=\mathbb{R}_{+}\times\mathbb{R}$ and
\begin{equation}
\varphi_{z}(y)=\frac{y}{z_{1}}-z_{2}\mathbf{1},\qquad z=(z_{1},z_{2})\in%
\mathbb{R}_{+}\times\mathbb{R}.  \label{eq:scale&translate}
\end{equation}
A result concerning the correctness of the PX-DA method when
$f_{Z}$ is improper is now stated. Although this has been
established explicitly in the case of either scaling or translation only
\citep{LiW99,MeV99}, the extension to the present setting is not
difficult. (See also Proposition 3 of \citet{HoM08}.)

\begin{proposition}
\label{prop:scaleAndtranslate}Consider the transformation in (\ref%
{eq:scale&translate}) and suppose that \newline
$c(y):=\int_{\mathbb{R}_{+}\times\mathbb{R}}f_{Y}(%
\varphi_{z}(y))J_{z}(y)dz_{1}dz_{2}$, $y\in \mathbb{R}^{q}$ is positive and
finite almost everywhere. Then the Markov
transition density on $\mathbb{R}^{q}$ defined by
\begin{equation*}
Q(y,B)=\int_{\mathbb{R}_{+}\times\mathbb{R}}\mathbb{I}_{B}(\varphi _{z}(y))%
\frac{f_{Y}(\varphi_{z}(y))J_{z}(y)}{c(y)}dz_{1}dz_{2}
\end{equation*}
is reversible with respect to $f_{Y}$.\
\end{proposition}

(Proof is in the Appendix.)

For any $h:\mathbb{R}^{p}\rightarrow\mathbb{R}$ which is square-integrable
with respect to $f_{X}$, i.e. $\int h^{2}(x)f_{X}(x)dx<\infty$, if a central
limit theorem holds, then
\begin{equation}
\frac{1}{\sqrt{L}}\sum_{n=1}^{L}h(X_{n})\overset{d}{\rightarrow}\mathcal{N}(%
\mathbb{E}_{f_{X}}(h(X)),\sigma^{2}(h))  \label{eq:clt}
\end{equation}
where
\begin{align}
\sigma^{2}(h) & =c_{0}(h)+2\sum_{i=1}^{\infty}c_{i}(h), &
c_{i}(h) & =\mathbb{E}_{f_{X}}(h(X_{i})h(X_{0}))-\mathbb{E}%
_{f_{X}}(h(X_{0}))^{2},\quad i\geq0.  \label{eq:clt_var}
\end{align}
The convergence in (\ref{eq:clt})\ is in distribution and the expectations
in the expression for $\sigma^{2}(h)$ are computed with respect to the law
of the Markov chain $\{X_{n}\}_{n\geq0}$ with initial distribution $f_{X}$.

\citet{HoM08} studied the relative performance of DA and PX-DA algorithms, 
addressing the case of Haar PX-DA - a class of algorithms involving
transformations derived 
from a particular group structure. The full details of Haar PX-DA are beyond the
scope of this article, but we note, 
for example, that if Algorithm \ref{alg:genPXDA} is modified to perform only a
scaling transformation and not translation, 
then it is an instance of Haar PX-DA (the reader is also directed to
\citet{LiW99} for details of the group structure underlying scaling and
translations).

The following inequality for the asymptotic variance of DA, PX-DA (for any
proper prior for $Z$) and Haar PX-DA is due to
\citet{HoM08},%
\begin{equation*}
\sigma ^{2}(h)\geq \sigma _{\text{P}}^{2}(h)\geq \sigma _{\text{H}}^{2}(h)
\end{equation*}%
where the subscripts indicate the algorithm generating $\{X_{n}\}_{n\geq 0}$;
the standard DA is without subscript, the subscript P denotes
PX-DA with a proper prior on $Z$ and H denotes Haar PX-DA.  Haar PX-DA is said
to be the
most efficient since it has a smallest asymptotic variance as
measured by (\ref{eq:clt}). It should also be noted that \citet{roy2012spectral}
has recently established that a general ``sandwich'' data augmentation algorithm
always converges as fast as its standard DA counterpart. Subject to the
transition kernel $Q$ in Proposition \ref{prop:scaleAndtranslate} being well
defined, it is immediate that our algorithm using that $Q$ is a sandwich data
augmentation algorithm, and thus enjoys this ordering property.

We stress that this variance inequality has only been shown to hold when
$f_{Y|X}$\ can be
sampled from exactly in step 1 of Algorithm \ref{alg:genPXDA}. In our numerical
experiments, this step is performed using a Metropolis-Hastings kernel, but as
we shall see, empirical results suggest that with this modification the PX-DA
algorithm still out-performs standard DA.

\section{A PX-DA sampler for the MDP model}
\label{sec:implementation}

The transformation of the augmented data will be as in (\ref%
{eq:scale&translate}). This section completes the description by specifying
the prior for the optimal value function, the auxiliary variable $Z$ and
culminates
with a statement of the complete sampling algorithm for these specific
choices. Extensions to a more general reward function and the practicality
of the approach for large problem sizes, specifically large $\mathcal{X}$,
are discussed at the end of the section.

Regarding the prior for $V$, the following requirements seem reasonable: (a)
it should respect the symmetry of the model regarding the $N$ states;
specifically, it should be invariant with respect to permutation of the
state labels; (b) it should be conjugate, so that Gibbs steps can be
implemented; and (c) to ease interpretation of the output, it should make
the model identifiable. These requirements are met by the following prior
distribution: a Gaussian $\mathcal{N} (\mathbf{0}_{N},\kappa I_{N})$
distribution (where $\kappa$ is a fixed hyper-parameter), 
but conditional on the event $\sum_{i=1}^N V(i)=0$. This prior
distribution may be alternately described as follows: take $U\sim\mathcal{N}
(\mathbf{0}_{N},\kappa I_{N})$, then set $V=U-N^{-1}\mathbf{1}_{N}\mathbf{1}%
_{N}^{\text{T}}U$, that is, remove the mean of the $U(i)$ to force the
components of $V$ to sum to zero.

The constraint $\sum_{i=1}^N V(i)=0$ addresses the additive
unidentifiability of the model, i.e. the fact that the likelihood is
unchanged if the same constant is added to all the $V(i)$. To fix
multiplicative unidentifiability, i.e. the likelihood is unchanged if
both $%
V $ and $\Sigma$ are multiplied by the same constant, we take
$\Sigma=I$ for the remainder of this Section. This choice presents an
important advantage: it makes it possible to implement Step 1 of
Algorithm 1 using an efficient Metropolis-Hastings step, as described
below. In Section \ref{sec:extensions}, we explain briefly how to
consider a more general matrix $\Sigma$, and why we believe that
$\Sigma=I$ should be sufficient in many practical (MDP) applications.

We note in passing a different way to treat additive
unidentifiability inspired by multivariate probit models
\citep{McR94}: i.e. set one of the $N$ components of the value
function to zero, e.g. $V(N)=0$. In our context however, this
would suppress the symmetry between the $N$ states, complicate the
notations, and bring no obvious benefit. Also, additive
unidentifiability can be exploited to yield a better PX-DA
sampler.

The augmented data is
\begin{equation*}
Y=(W_{1},\ldots,W_{T}),\text{ with }p(\left.
w_{1},\ldots,w_{T}\right\vert v)=\prod\limits_{i=1}^{T}p(\left.
w_{i}\right\vert
v)=\prod\limits_{i=1}^{T}\mathcal{N}(w_{i};R_{i}v,I_{M}),
\end{equation*}
which may be viewed as arising from ``disintegration'' of \eqref{eq:ccp}.
Indeed, the PX-DA algorithm defined below will be derived from the joint density
($%
f_{X,Y}(x,y)$ in section \ref{sec:pxda}, with $x= v$ and $y=
w_{1},\ldots,w_{T}$):
\begin{align}
& p(v,w_{1},\ldots,w_{T}|d)  \notag \\
& \propto\mathcal{N}\left( v;\mathbf{0}_{N-1},\kappa I_{N-1}-\kappa N^{-1}%
\mathbf{1}_{N-1}\mathbf{1}_{N-1}^{\text{T}}\right) \prod\limits_{i=1}^{T}%
\mathbb{I}_{\{w_{i}\in\mathbb{R}^{M}:\,w_{i}(a_{i})\geq w_{i}(j),j\neq
a_{i}\}}\mathcal{N}(w_{i};R_{i}v,I_{M}),  \label{eq:targetDA}
\end{align}
with the slight abuse of notation that the vector $v$ in $\mathcal{N}%
(w_{i};R_{i}v,I_{M})$ is $N$ dimensional where $N$-th component is
\begin{equation*}
v(N)=-\sum_{i=1}^{N-1}v(i).
\end{equation*}
(This convention will hold for the remainder of this section wherever $%
R_{i}v $ occurs.) The density (\ref{eq:targetDA}) clearly admits the posterior
over $v$ as a marginal. Direct simulation from $p(w_1,...,w_T|v,d)$ is difficult
in general, due to the presence of truncated Gaussian distributions. This is
where our Metropolis-Hastings step will come in; further discussion is postponed
until the end this section. Putting aside this difficulty for now, we next
describe a PX-DA algorithm for this model, i.e. derived from the standard DA
algorithm which iteratively samples from the conditionals $p(v|w_1,...,w_T,d)$
and $p(w_1,...,w_T|v,d)$.

The transformation of the augmented data for the PX-DA scheme is
given in (\ref{eq:scale&translate}). We set $%
f_{Z_{1},Z_{2}}(z_{1},z_{2})=f_{Z_{2}}(z_{2})f_{Z_{1}}(z_{1})$ and
\begin{equation}
Z_{1}\sim IG(a,b),\quad Z_{2}\sim\mathcal{N}(0,\kappa/N),
\label{eq:workingParams}
\end{equation}
where $IG$\ is the inverse Gamma density. We stress that here $\kappa$ is the
same parameter as appearing in the prior distribution over $U$, specified
earlier in this section. It is this structure which allows us to construct a
PX-DA algorithm incorporating a translation move.

 To clarify the connection with the
description of the generic PX-DA sampler in section \ref{sec:pxda}, with a
slight abuse of the definition of $\varphi_{z}^{-1}$,
\begin{equation*}
Y^{\prime}=\varphi_{z}^{-1}(Y)=\left( \sqrt{z_{1}}(W_{1}^{\text{T}}+z_{2}%
\mathbf{1}_{M}^{\text{T}}),\ldots,\sqrt{z_{1}}(W_{T}^{\text{T}}+z_{2}\mathbf{%
1}_{M}^{\text{T}})\right) ^{\text{T}},
\end{equation*}
and the Jacobian in (\ref{eq:jacobian}) is%
\begin{equation*}
J_{z}(y^{\prime})=z_{1}^{-\frac{MT}{2}}.
\end{equation*}
With this choice of transformation of the variables, step 1 and 2a of the
generic PX-DA algorithm \ref{alg:genPXDA} can be combined into step 1 of
algorithm \ref{alg:inferValue} below. Similarly, step 2b and 3 of algorithm %
\ref{alg:genPXDA} may be combined into step 2 of algorithm \ref%
{alg:inferValue}.

The Metropolis Hastings kernel (with independent proposals) for
step 1 of Algorithm \ref{alg:inferValue} presented in the Appendix
is quite efficient with acceptance rates typically around 70
percent for the numerical examples in Section
\ref{sec:experiments}. Step 2 can be implemented as detailed in
Section \ref{sec:algInferValStep2}. When improper priors are used for $V$, $%
Z_{1}$ and $Z_{2}$, the corresponding terms in
(\ref{eq:step2Density}) should be omitted. As discussed in Section
\ref{sec:pxda}, when improper priors are used for $Z_1$ and $Z_2$,
these variables should not be sampled in step 1 above. However, one should be 
careful that (\ref{eq:targetDA}) is still well defined when $\kappa=\infty$
otherwise $\kappa$ should always
be set to a finite value.
(For instance, if the observed process is constant, then $\kappa=\infty$ gives 
an improper posterior. However, we have been able to establish that $\kappa=\infty$ can
give a proper posterior under quite general conditions; details may be obtained from the authors.)

\subsection{Extensions\label{sec:extensions}}

\subsubsection{Action dependent Rewards}

In Section \ref{sec:problemState} it was assumed that the reward function is
not action dependent. The following extension to the criterion in (\ref%
{eq:valueFnForMu}) can be considered. Replace $r(X_{t})$ in (\ref%
{eq:valueFnForMu})\ by
\begin{equation}
r(X_{t},A_{t})=r_{1}(X_{t})+r_{2}(A_{t}).  \label{eq:additivereward}
\end{equation}%
\noindent\hrulefill

\begin{algorithm}
\textbf{\label{alg:inferValue} PX-DA for the MDP model}
\end{algorithm}

Let $w_{1:T}$\ and $v$ be the samples after iteration $n$. At iteration $n+1$%
, perform the following two steps.

\textbf{Step 1:} Sample $Z_{1}\sim IG(a,b)$, call the result
$z_{1}$, sample $Z_{2}\sim\mathcal{N}(0,\kappa/N)$ and let $z_{2}$
denote this sampled value. For each $i=1,...,T$, sample $W_{i}$\
from the truncated Gaussian
\begin{equation}
\mathbb{I}_{\{w_{i}\in\mathbb{R}^{M}:\,w_{i}(a_{i})\geq
w_{i}(j),j\neq a_{i}\}}\mathcal{N}(w_{i};R_{i}v,I_{M}),
\label{eq:step1Density}
\end{equation}
call the result $w_{i}$ and set $w_{i}^{\prime}=\sqrt{z_{1}}(w_{i}+z_{2}%
\mathbf{1}_{M})$. (This step can be achieved directly or using the
Metropolis-Hastings kernel detailed in Section
\ref{sec:MHkernel}.)

\textbf{Step 2:} Sample $(V(1),\ldots,V(N-1),Z_{2},Z_{1})$ from
the joint
density%
\begin{align}
& \mathcal{N}\left( v;\mathbf{0}_{N-1},\kappa I_{N-1}-\kappa N^{-1}\mathbf{1}%
_{N-1}\mathbf{1}_{N-1}^{\text{T}}\right) \prod \limits_{i=1}^{T}\mathcal{N}%
\left(\frac{w_{i}^{\prime}}{\sqrt{z_{1}}}-z_{2}\mathbf{1}_{M};R_{i}v,I_{M}%
\right)  \notag \\
&
\times\mathcal{N}(z_{2};0,\kappa/N)z_{1}^{-\frac{MT}{2}}IG(z_{1};a,b)
\label{eq:step2Density}
\end{align}
and $z_{1}^{-0.5}w_{i}^{\prime}-z_{2}\mathbf{1}_{M}$,
$i=1,$\ldots$,T$, are now the final $w_{1:T}$ for iteration $n+1$.

\noindent\hrulefill

Note that $V$ for this new problem still satisfies 
\eqref{eq:optValueFn} with the reward function therein replaced by
\eqref{eq:additivereward}. In this case the action generation model is now
\begin{equation*}
A_{t}=\arg \max_{a\in \mathcal{A}}\left\{ \epsilon _{t}(a)+r_{2}(a)+\beta
\left(R_t V \right)(a)
\right\}
\end{equation*}
and
\begin{equation*}
p(A_{t}=i\left\vert v,r_{2},\Sigma ,x_{t}\right. )=\int_{\{u \in
\mathbb{R}^{M}:\, u(i)\geq u(j)\mbox {for all }j\neq i\}}
\mathcal{N}(u ;r_{2}+R_{t}v,\Sigma )du.
\end{equation*}
It can be verified that, for all $(z_{1},z_{2},z_{3})\in \mathbb{R}%
_{+}\times \mathbb{R}\times \mathbb{R}$,
\begin{equation*}
p(A_{t}=i\left\vert \sqrt{z_{1}}(v+z_{2}\mathbf{1}),\sqrt{z_{1}}(r_{2}+z_{3}%
\mathbf{1}),z_{1}\Sigma ,x_{t}\right. )=p(A_{t}=i\left\vert v,r_{2},\Sigma
,x_{t}\right. ).
\end{equation*}%
The prior for $V$ could be the same as before (see Section \ref%
{sec:implementation}) and one could also use a prior with the same structure
for $r_{2}$.

\subsubsection{Large State-spaces}

Since $V$ is a vector of length $|\mathcal{X}|$, the approach
detailed thus far will be impractical for a very large state-space
$\mathcal{X}$. In this setting we may regress the optimal value function
onto a set of basis functions. (A similar approach was proposed by
\citet{GeK96}, \citet{GeK00} for a finite horizon dynamic discrete
choice problem and the idea goes back some way in the control literature, see
for example \citet{Sch85}) Let $\{\phi _{i}\}_{1\leq i\leq K}$ be a
collection of basis functions, mapping $\mathcal{X}$ to the real
line. Typically $K$ is much smaller than $\left\vert
\mathcal{X}\right\vert $. It is assumed that the conditional
expectation, $\sum_{x^{\prime }\in \mathcal{X}}\phi _{i}(x^{\prime
})p(x^{\prime }|x,a),$ can be computed easily for each state-action pair $%
(x,a)$ and $i$. For example, this would be true if $p(x^{\prime
}|x,a)$\ is non-zero for only a handful of values of $x^{\prime
}$, see the human controller example considered in Section
\ref{sec:tetris_results}. The
action generation model is (for an action independent reward),%
\begin{equation*}
a_{t}=\arg \max_{a\in \mathcal{A}}\left\{ \epsilon
_{t}(a)+\sum_{i=1}^{K}V(i)\left(R_t \phi_i \right)(a)
\right\} ,
\end{equation*}%
and the corresponding likelihood satisfies%
\begin{equation*}
p(A_{t}=i\left\vert \sqrt{z_{1}}v,z_{1}\Sigma ,x_{t}\right.
)=p(A_{t}=i\left\vert v,\Sigma ,x_{t}\right. ).
\end{equation*}%
The likelihood is no longer invariant to scalar translations of the value
function. As the model is no longer additively unidentifiable, an
unconstrained prior may thus be defined over all $N$ components of $V$. For
example, the prior $N(\mathbf{0}_{N},\kappa I_{N})$\ is admissible even as $%
\kappa \rightarrow \infty $. The PX-DA implementation for this model will
involve transforming the augmented data by a scalar multiplication only.

\subsubsection{Constrained Actions}

In some applications, state dependent action constraints are
present, i.e. not every action in $\mathcal{A}$\ is permitted in
every state. The modification to Algorithm \ref{alg:inferValue} is
trivial. For example, if action $j$ is not permitted in state
$x_{i}$, then row $j$ of the $R_{x_{i}}$ defined in
(\ref{eq:matrixR}) is deleted. Action constraints are present in
the example studied in Section \ref{sec:experiments}.

\subsubsection{Non-identity Noise Covariance Matrix}

We think that restricting the model to an identity covariance matrix
for the noise term is very reasonable for the
following reasons. First, in the MDP context, one is mostly interested
in inferring $V$ as $\Sigma$ is merely a nuisance parameter. Second,
since only one action is observed at a time, it seems hard to estimate
correlations between the different components of the noise
vector. Third, considering a general $\Sigma$ means that the dimension
of the parameter space becomes $O(M^2)$, and the computational burden
$O(M^3)$, as opposed to $O(M)$ for both quantities in the $\Sigma=I$
case. (The computational burden increases also because of the greater
difficulty to sample the latent variables $W_i$, as explained below.)
This is clearly impractical when $M$ is large.

However, for the sake of completeness, we now explain how to
account for a general covariance matrix $\Sigma $.  The prior
suggested in \citet{ImV05} may adapted to the present setting. A
prior for the covariance matrix $\Sigma $\ subject to the
constraint $[\Sigma ]_{1,1}=1$ is constructed by normalizing the
samples from an inverse Wishart distribution.  Specifically,
$\widetilde{\Sigma }\sim \mathcal{IW}(\nu ,\widetilde{S})$ and
$\Sigma =\widetilde{\Sigma
}/[\widetilde{\Sigma }]_{1,1}$. Let $z_{1}=[%
\widetilde{\Sigma }]_{1,1}$ then,%
\begin{equation*}
p(z_{1},\Sigma )\propto |\Sigma |^{-(\nu +M+1)/2}\exp \left( -\frac{\alpha
^{2}}{2z_{1}}\text{tr}(S\Sigma ^{-1})\right) (z_{1})^{-\frac{\nu M}{2}-1}
\end{equation*}%
where constant $\alpha ^{2}$\ satisfies $\widetilde{S}=\alpha ^{2}S$. The
conditional density
\begin{equation*}
p(z_{1}|\Sigma )=IG\left( \frac{\nu M}{2},\frac{\alpha ^{2}}{2}\text{tr}%
(S\Sigma ^{-1})\right)
\end{equation*}%
is now the new distribution for the scaling parameter in the PX-DA
transformation of the augmented data; see (\ref{eq:workingParams}). To infer
$\Sigma $\ as well, Algorithm \ref{alg:inferValue} would be modified to
sample $W_{1:T}$, $V$ and then $\Sigma $ in turn. For a non-diagonal
covariance matrix, step 1 cannot be implemented with the Metropolis-Hastings
kernel described in Section \ref{sec:MHkernel}. A possible alternative is to
use a Gibbs sampling step where, for each $i$, each component of $W_{i}$\ is
sampled conditioned on the remaining components. Once a complete cycle has
been performed, then the transformation at the end of step 1 can be applied.
Step 2 will be modified to sample $(Z_{1},Z_{2},V,\Sigma )$ conditioned on $%
w_{1:T}$, which can be performed by an appropriate blocking scheme after the
change of variable in (\ref{eq:changeOfVar}). Roughly speaking, $%
(Z_{1},Z_{2},V)$ is sampled conditioned on $\left( \Sigma ,w_{1:T}\right) $\
and then $(Z_{1},\Sigma )$ conditioned on $\left( Z_{2},V,w_{1:T}\right) $.
The samples produced may suffer from much more correlation than in the case
of Algorithm \ref{alg:inferValue} which is catered to $\Sigma =I$.

\section{Numerical Examples}
\label{sec:experiments}

\subsection{Toy Example}

To demonstrate the performance improvements of PX-DA over standard DA, a
data record of $20$ state-action pairs was generated from the model with 7
states, 3 actions. The true optimal value function was drawn from the prior.
Algorithm \ref{alg:inferValue} was run for $5\times10^{5}$ iterations and
half were discarded for burn in. The parameters of the priors in (\ref%
{eq:workingParams}) were chosen $a=b=1$, $\kappa=2500$. Figure \ref%
{fig:additveImprovements} shows the empirical auto-correlation of the MCMC
output for some of the
components of the estimated optimal value function. The improvements due to
scaling
and translation of the augmented data are isolated. For the components of
the value function not shown, the improvements were comparable.
The acceptance rate for the Metropolis-Hastings\ kernel used to implement step 1
of Algorithm \ref%
{alg:inferValue} was in excess of $95\%$.

\begin{figure}[h!]
\centering
\includegraphics[width=0.49%
\textwidth]{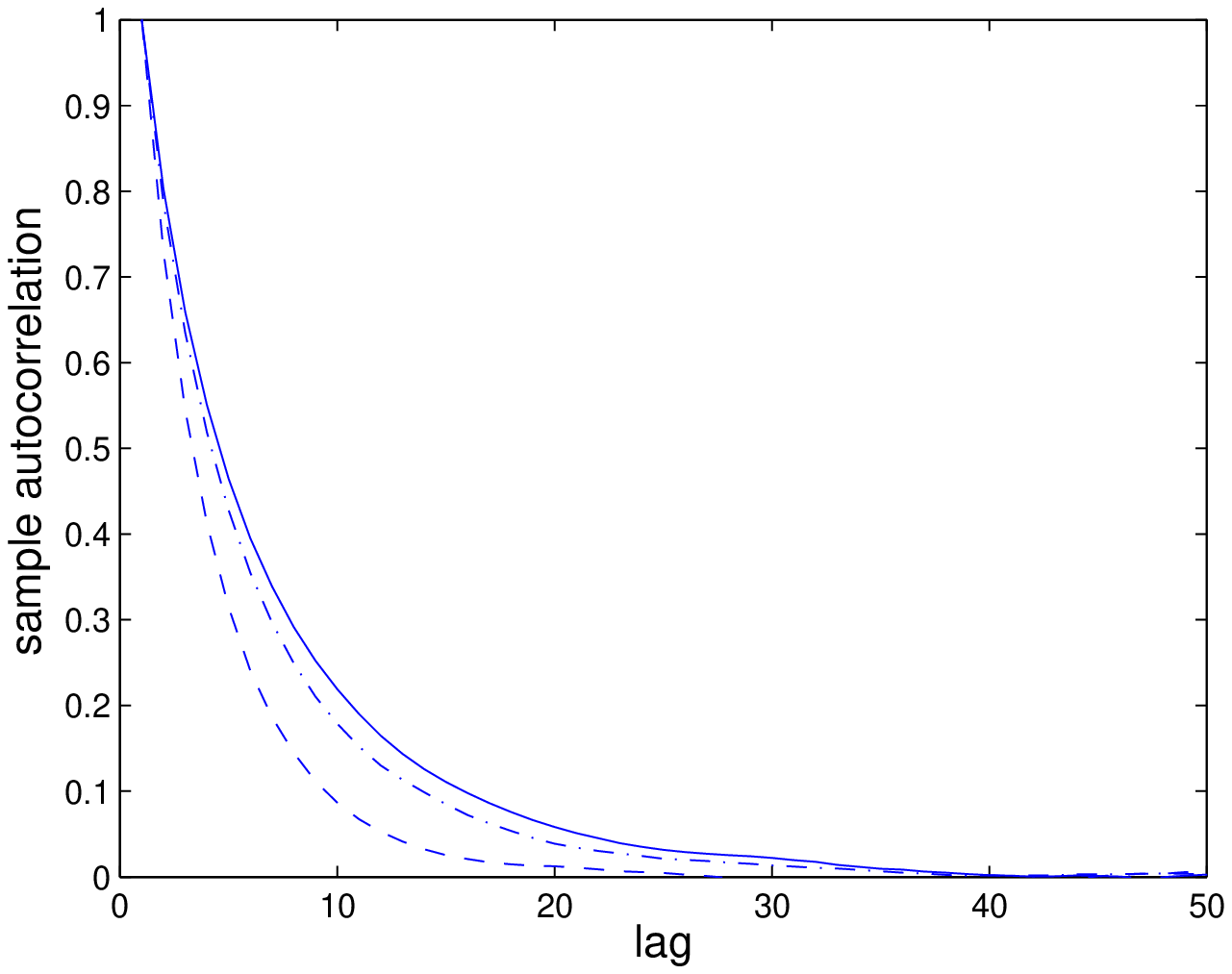} %
\includegraphics[width=0.49%
\textwidth]{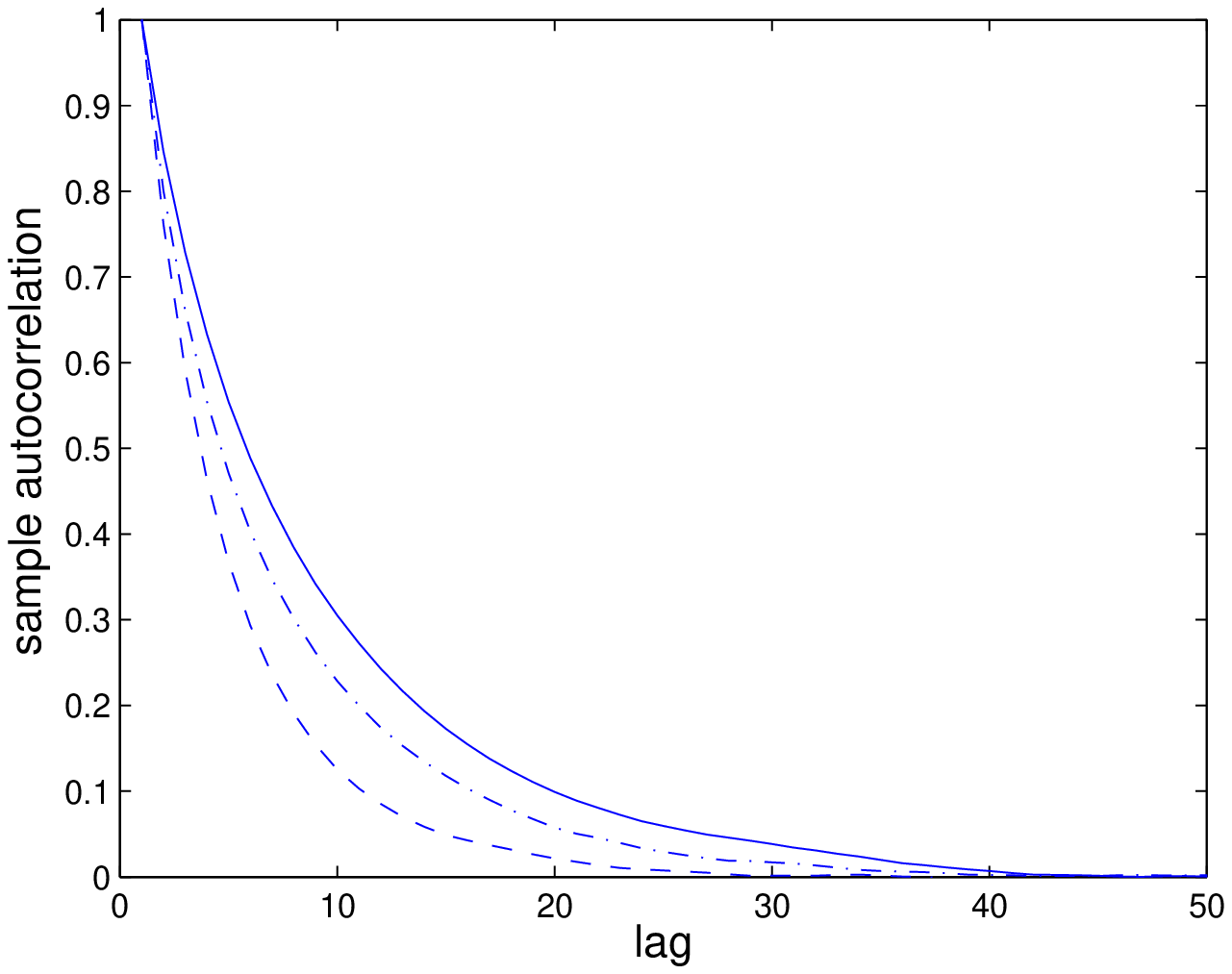}
\caption{Post-burn in autocorrelation plots of
posterior samples for components (left) 4 and (right) 7 of the value function.
Solid
line is standard DA, dash-dot is PX-DA with scale move only, dashed line is
PX-DA with scale and translation move.}
\label{fig:additveImprovements}
\end{figure}

\begin{figure}[h!]
\centering
\includegraphics[width=0.49%
\textwidth]{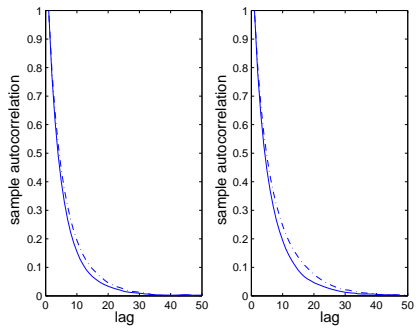}
\caption{PX-DA post-burn in autocorrelation plots of posterior samples for
component 4 (left) and 7 (right) of the value function. Dash-dot line is
PX-DA with scale move only and a proper prior with $a=5$, $b=0.5$. Solid
line is PX-DA with scale move only and an improper prior.}
\label{fig:improperImprovements}
\end{figure}

Figure \ref{fig:improperImprovements} isolates the effect of an improper
prior in PX-DA (i.e. $a=b=0$, $\kappa=\infty$). This study is restricted to PX-DA that scales the augmented
data only since the prior for the value function in (\ref{eq:targetDA})
also depends on the parameter $\kappa$\ that controls the variance of the
law of the translation parameter (\ref{eq:workingParams}). Figure \ref%
{fig:improperImprovements} shows the computed autocorrelation for components
4 and 7 of the estimated value function. For the proper prior, $a=5$, $b=0.5$. 
In this case the improper prior for the scaling parameter yields a modest
improvement in performance over the proper prior. (Note though there no
longer the issue of tuning the prior for the scaling parameter.)

\begin{figure}[h!]
\centering
\includegraphics[width=0.49%
\textwidth]{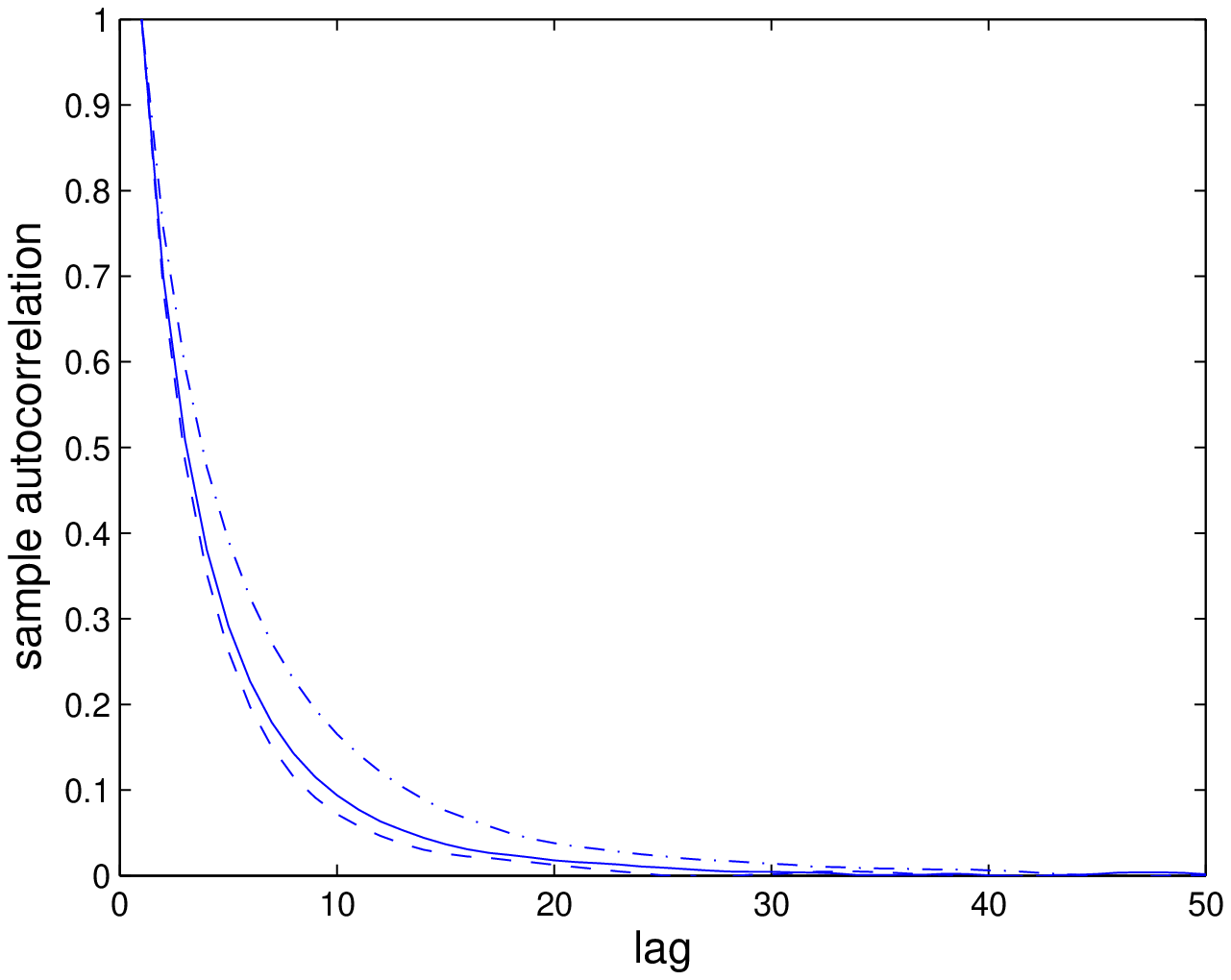} %
\includegraphics[width=0.49%
\textwidth]{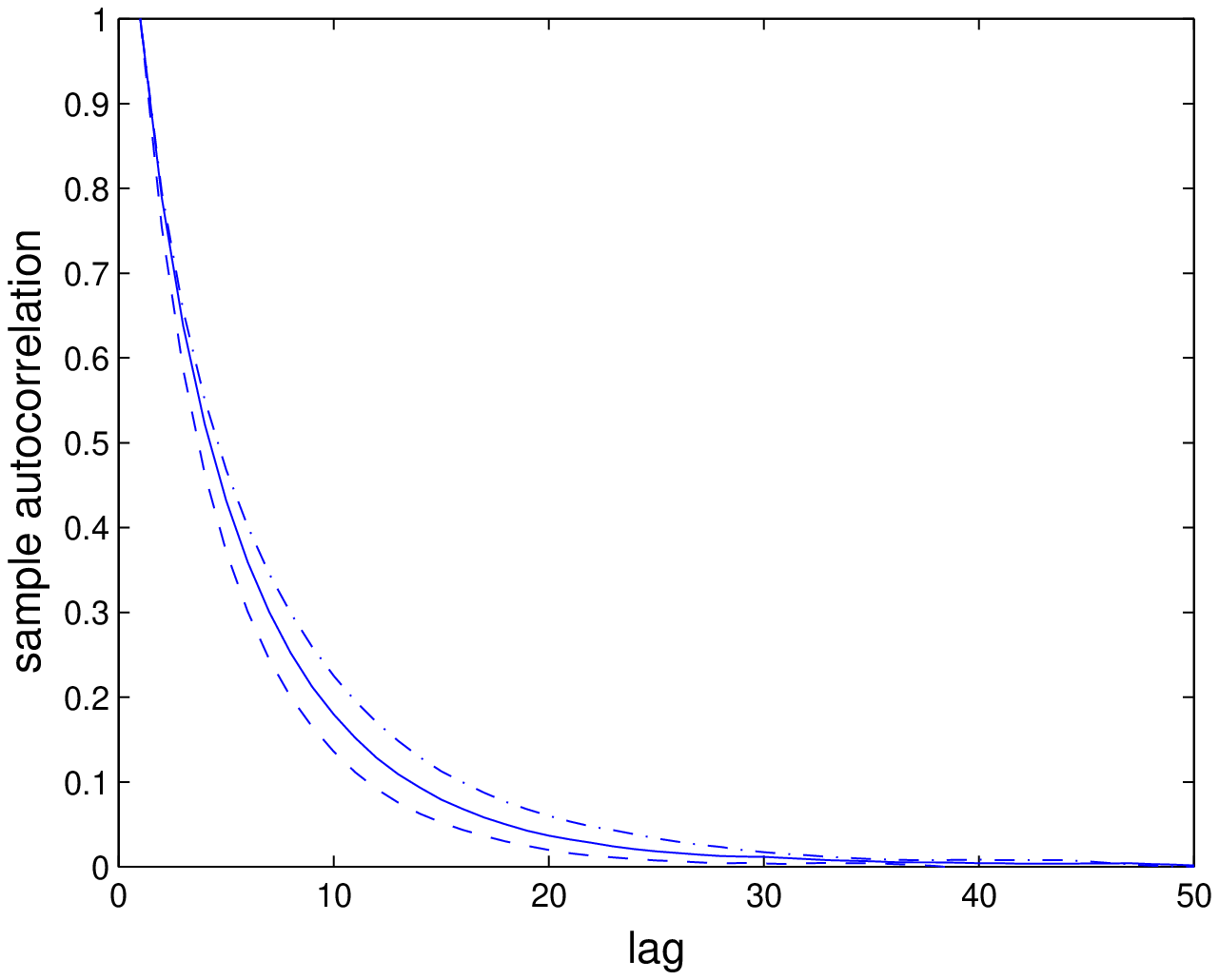}%
\newline
\includegraphics[width=0.49%
\textwidth]{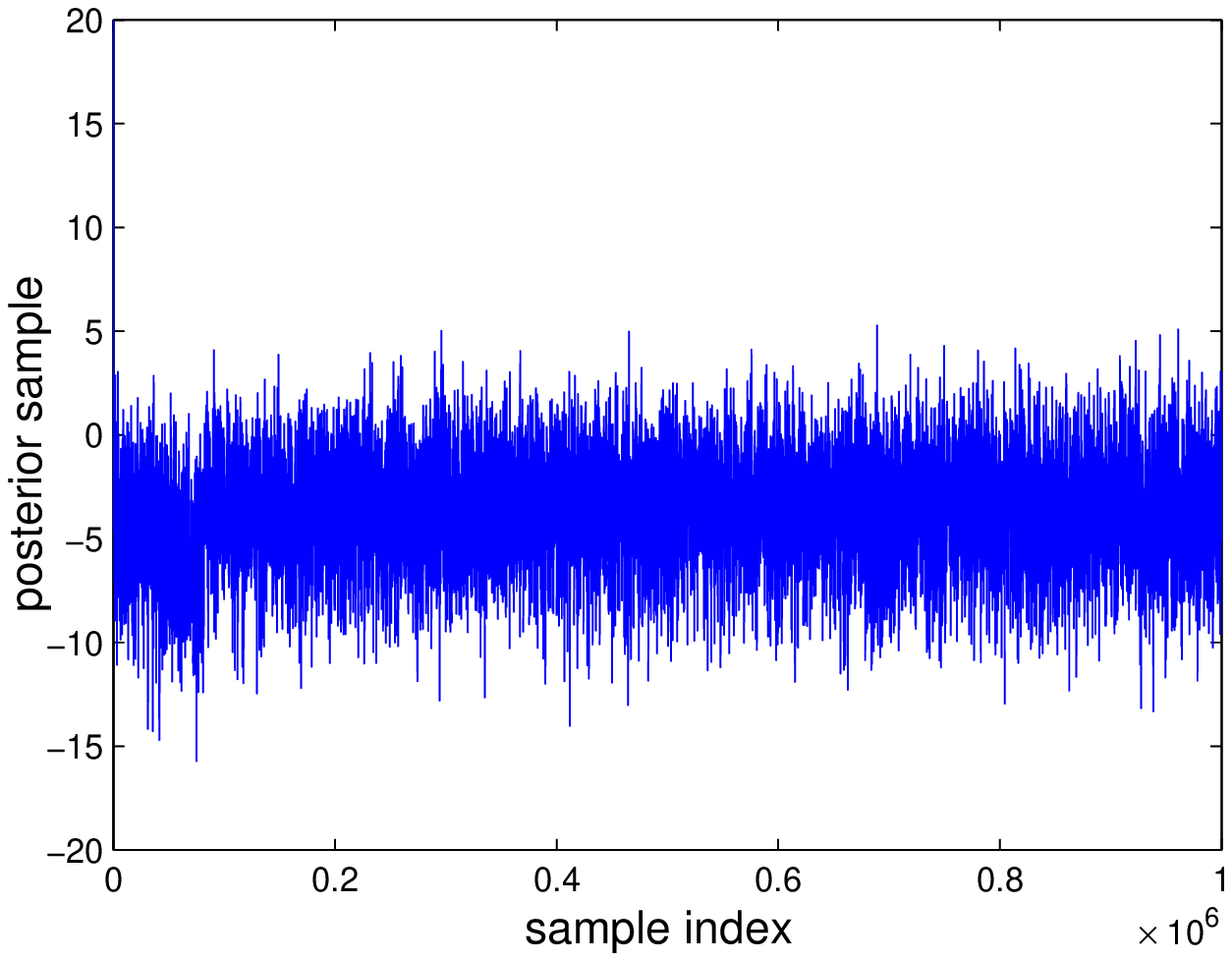}
\caption{From left to right: PX-DA post-burn in autocorrelation plots of
posterior samples for component 4 and 7 of the value function; and DA trace
plot of posterior samples of component 7 of the value function, lack of
stationarity is apparent. For the
autocorrelation plots, solid line is PX-DA with translation move only,
dash-dot is PX-DA with scale move only, dashed line is PX-DA with scale and
translation move. The mean of the second half of the samples from DA is $%
-3.56$ whereas the true posterior mean (calculated with PX-DA for which the
three implementations are in agreement) is $-8.4.$. It appears the DA algorithm
has not converged. }
\label{fig:nonConvergence}
\end{figure}

The final experiment demonstrates a situation where PX-DA converges but DA
does not. The data comprising of $20$ state-action pairs of the previous
examples is extended to $50$ by appending $30$ more. In this example, the
priors for $V$, $Z_{1}$ and $Z_{2}$ are improper. The posterior mean of the
value function calculated with PX-DA with scaling and translation is $%
[2.34,-0.92,-1.02,4.12,5.69,-1.79,-8.41]^{\text{T}}$. ($1.5\times 10^{6}$
posterior samples but half discarded for burn in. The posterior mean for
PX-DA with scaling or translation only was practically the same.) The PX-DA
implementations were initialized with the value function set to $100\times
\mathbf{1}_{7}$. Shown in Figure \ref{fig:nonConvergence} is the trace plot
of the samples of component 7 of the value function obtained using the DA
method initialized with the value function set to $10\times \mathbf{1}_{7}$.
The mean of the second half of the samples in Figure \ref{fig:nonConvergence}
is $-3.56$. (In fact all other components of the mean of the posterior value
function calculated with DA are quite far out.) In this case we see that DA
fails to converge even though initialized far closer to the true values than
PX-DA. Finally, to isolate the improvements due to scaling and translation,
the autocorrelation plots of certain components of the posterior samples of
the value function are compared in Figure \ref{fig:nonConvergence} for PX-DA
implemented with both additive and scaling, scaling only and additive only.
In this example, the translation move appears more beneficial than scaling.

\subsection{Application to Human Controller Learning}

\label{sec:tetris_results}

In this section we apply the proposed method to an MDP which arises
in the context of the popular computer game Tetris. In this game
the player controls the positions and orientations of random
two-dimensional shapes, henceforth the \emph{blocks}, which arrive
over time and occupy a field of play, henceforth the \emph{board},
in a non-overlapping manner.


\begin{figure}[h!]
\begin{center}
\includegraphics[width=0.7\textwidth]{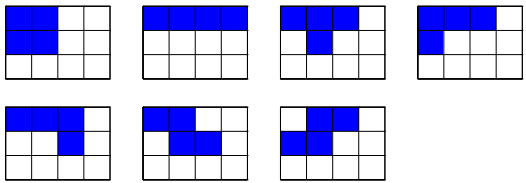}
\end{center}
\caption{The seven blocks of Tetris.}
\label{fig:tetris_blocks}
\end{figure}
\begin{figure}[h]
\begin{center}
\includegraphics[width=0.7\textwidth]{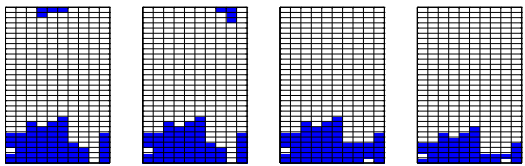}
\end{center}
\caption{Example iteration of Tetris. From left to right: 1) A
block appears at the top of the board. 2) An action is taken to
rotate and translate the block. 3) The block then falls until it
reaches occupied squares. 4) Fully occupied rows of the board are
removed.} \label{fig:tetris_board_demo}
\end{figure}

\subsubsection{Model Definition}

In the MDP formulation of Tetris, the state $X=(\zeta ,\eta)$
consists of two components. The first component, $\zeta$, is the
current configuration of the board and is expressed as a $30\times
10$ binary matrix. The second component, $\eta$, is the index of a
block. We consider $7$ distinct blocks, shown in Figure
\ref{fig:tetris_blocks}, and thus $\eta$ takes values in
$\{1,2,...,7\}$. Each action $A$ consists of the angle through
which to rotate the current block ($0^{\circ}$, $90^{\circ}$,
$180^{\circ}$, $270^{\circ}$), and the number of squares by which
to move it left or right. For each state not all combinations of
horizontal translation and rotation are necessarily permitted as
the block must remain entirely within the boundaries of the board
and must not overlap with any occupied squares. We write
$\mathcal{A}(x)$ for the set of actions which are valid in state
$X=x$.

From the current state $X_{t}=(\zeta_{t},\eta_{t})$ and an action $A_{t}\in%
\mathcal{A}(X_{t})$, the evolution of the state occurs according to
\begin{align}
\zeta_{t+1}  =\psi(\zeta_{t},\eta_{t},A_{t}), & &  \eta_{t+1}
\sim\mathcal{U}(1,2,...,7),  \label{eq:tetris_mdp_dynamics}
\end{align}
where $\psi$ is a deterministic mapping which describes the evolution of the
board configuration once the action has been chosen. For a configuration $%
\zeta_{t}$\ with no occupied squares in the top row, $\psi$ yields the new
configuration $\zeta_{t+1}$ by moving the block $\eta_{t}$ according to $%
A_{t}$, then allowing the block to ``fall'' until it reaches an 
occupied square or the bottom row of the board, and then removing
any fully occupied rows. For a configuration $\zeta_{t}$
which has an occupied square in the top row, $\psi$ sets $%
\zeta_{t+1}=\zeta_{t}$ irrespective of $A_{t}$ and $\eta_{t}$. The
latter corresponds to \textquotedblleft
termination\textquotedblright\ of the game; once such a state is
reached, subsequent actions do not influence the state. A
pictorial representation of one iteration of the game is given in
Figure \ref{fig:tetris_board_demo}.

As each state consists of a single board configuration and block
type, the total number of states in the Tetris model is rather
large. We therefore adopt the approach outlined in Section
\ref{sec:implementation} and regress the value
function on to a collection of $K$ basis functions $\{\phi_{1},\phi_{2},...,%
\phi_{K}\}$, which depend on the board configuration $\zeta$ but not on the
randomly falling piece $\eta$. (Note that the latter is not controlled as
new blocks arrive independently of the action and the previous state.)
Specific details of the basis functions are given in section
\ref{sec:tetris_results}.

We assume that the reward is independent of the action and the action
generation model is then
\begin{equation}
A_{t}=\arg \max_{a\in \mathcal{A}(x_{t})}\left\{ \epsilon
_{t}(a)+\sum_{i=1}^{K}V(i)\cdot \left( \phi _{i}\circ \psi \right)
(\zeta _{t},\eta _{t},a)\right\} .  \label{eq:tetris_obs_gen}
\end{equation}%
We assume that the noise corrupting the action choice has identity
covariance. The likelihood of observed data is
\begin{equation*}
p(d|v,\Sigma )=\prod_{t=1}^{T}p(a_{t}|v,x_{t},\Sigma ),
\end{equation*}%
where for each $k=1,...,T$,
\begin{equation*}
p(A_{t}=i\left\vert v,x_{t},\Sigma \right. )=\int_{\{\epsilon \in \mathbb{R}%
^{M_{t}}:\,\epsilon (i)\geq \epsilon (j) \mbox{ for all }j\neq
i\}}\mathcal{N}(\epsilon
;R_{t}v,I_{M_{t}})d\epsilon .
\end{equation*}%
Here $M_{t}:=|\mathcal{A}(x_{t})|$ and $R_{t}$ is the $M_{t}\times N$ matrix
with entries specified by
\begin{equation*}
\lbrack R_{t}]_{ij}:=(\phi _{j}\circ \psi )(\zeta _{t},\eta _{t},i).
\end{equation*}%
In this case the likelihood is invariant to scaling in the sense that
\begin{equation*}
p(d|v,\sigma ^{2}I)=p(d|\sqrt{z_{1}}v,z_{1}\sigma ^{2}I),\;\;\;\forall
z_{1}\in \mathbb{R}_{+}.
\end{equation*}%
The sampling algorithm\ for inference is Algorithm
\ref{alg:inferValue} where the augmented data and transformation
are given by
\begin{align*}
Y=(W_{1},\ldots ,W_{T}),&&Y^{\prime }=\varphi _{z_{1}}^{-1}(Y)=\left(
\sqrt{z_{1}}W_{1}^{\text{T}%
},\ldots ,\sqrt{z_{1}}W_{T}^{\text{T}}\right) ^{\text{T}},
\end{align*}%
where the scaling factor $Z_{1}\sim IG(a,b)$. The Jacobian for
this transformation is
\begin{equation*}
J_{z_{1}}(y^{\prime })=z_{1}^{-\frac{\sum_{t=1}^{T}M_{t}}{2}}.
\end{equation*}%
The prior for $V$ is $\mathcal{N}(\mathbf{0}_{N},\kappa I_{N})$.

We consider the following $K=3$ basis functions which were found
to capture various features of the board configuration.
$\protect\phi_{1}$, the height of the top-most occupied square in
the board, across all the columns; $\protect\phi_{2}$, the number
of unoccupied squares which have at least one occupied square
above them in the same column; $\protect\phi_{3}$ the sum of the
squared differences between occupied heights of adjacent columns.

In \citet{TsV94}, \citet{BeT96}, for a board with $c$ columns, $\phi
_{1}$, $\phi _{2}$ and $2c-1$\ additional features were used to
construct an automated self-improving Tetris playing system using
Reinforcement Learning techniques. In contrast, the emphasis here
is to make predictions about actions and mimic play on the basis
of observed state-action data. In our setup the latter amounts to
posterior prediction, which can be performed in the following manner. Let $%
\{V_{n}\}_{n=1}^{L}$ be a collection of post-burn-in samples from the
posterior distribution over the value function, obtained from the PX-DA
algorithm. Then for each state in a given sequence $\{(\zeta _{t},\eta
_{t})\}_{t=1}^{T}$ we would like to make predictions under our model about
the corresponding action, on the basis of the posterior samples $%
\{V_{n}\}_{n=1}^{L}$. To this end, for each $(\zeta _{t},\eta _{t})$ we
define the MAP predicted action as
\begin{align}
\widehat{A}_{t}^{\text{MAP}}(\zeta _{t},\eta _{t})& :=\arg \max_{a\in
\mathcal{A}(x_{t})}\sum_{n=1}^{L}\mathbb{I}\left[ \widehat{A}_{n,k}(\zeta
_{t},\eta _{t})=a\right] ,  \notag \\
\widehat{A}_{n,k}(\zeta _{t},\eta _{t})& :=\arg \max_{a\in \mathcal{A}%
(x_{t})}\left\{ \epsilon _{n,k}(a)+\sum_{j=1}^{N}V_{n}(j)\cdot \left( \phi
_{j}\circ \psi \right) (\zeta _{t},\eta _{t},a)\right\} ,
\label{eq:MAP estimate}
\end{align}%
where for each $1\leq n\leq L$, $1\leq k\leq T$ and $a\in \mathcal{A}(x_{t})$%
, $\epsilon _{n,k}(a)$ is an independent $\mathcal{N}(0,1)$ random variable.

In the following section the predictive performance of the model is assessed
for a number of data sets. Each data set is divided into two subsets. The
PX-DA algorithm is used to draw samples from the posterior corresponding to
the first subset and then the accuracy of the posterior prediction is
assessed using the second subset. This assessment is performed in terms of
the empirical action error, defined as
\begin{equation}
\mathcal{E}_{a}:=\frac{1}{T}\sum_{t=1}^{T}\mathbb{I}\left[ \widehat{A}_{t}^{%
\text{MAP}}(\zeta _{t},\eta _{t})\neq a_{t}\right] .  \label{eq:error_def}
\end{equation}%
where $\{(\zeta _{t},\eta _{t},a_{t})\}_{t=1}^{T}$ is the second data subset.

Finally we note that in practical situations computation of
\eqref{eq:MAP
estimate} may be expensive if $L$ is large, in which case one may resort to
heuristic action prediction based on a posterior point estimate of $V$. We
do not explore this issue further.

\subsubsection{Experiment 1}

\begin{figure}[h!]
\centering
\includegraphics[width=0.49\textwidth]{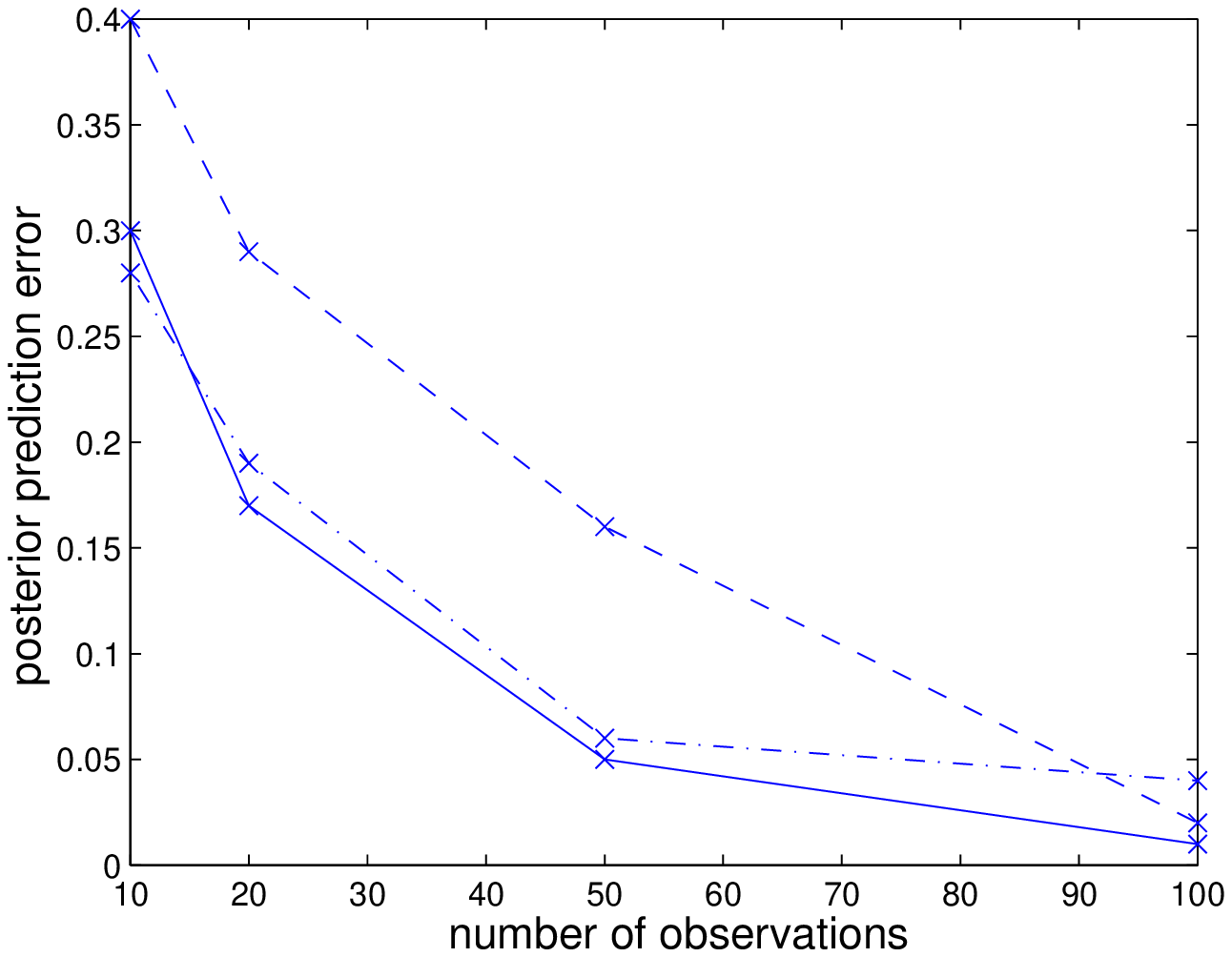} %
\includegraphics[width=0.49\textwidth]{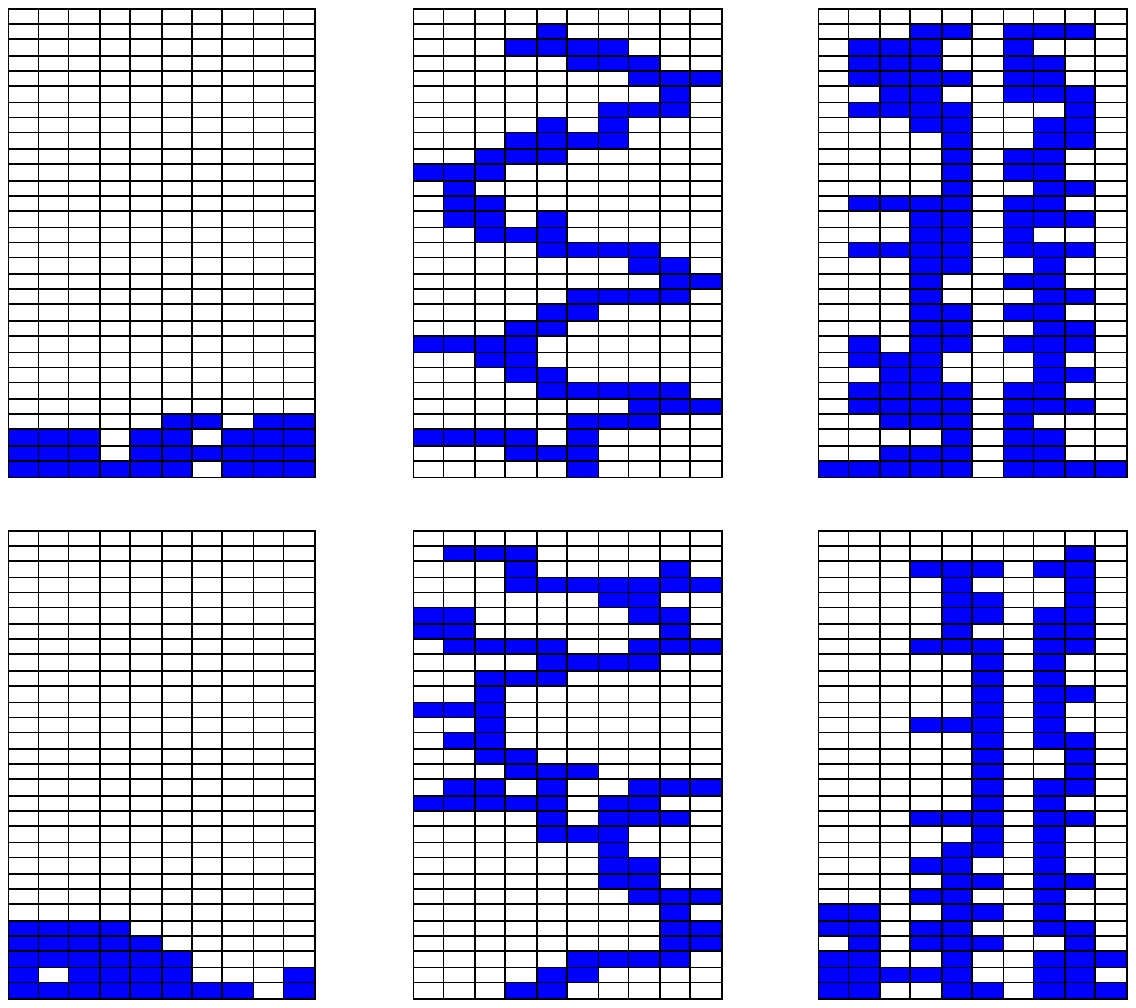}
\caption{Experiment $1$. Right top row: Board snap-shots.
From left to right the true value functions are $(-3, -15, -1)$, $(0, 5, 0)$
and $(-20, 0, 1)$. Right bottom row: Board snapshots during play according
to posterior predictive actions for a different block sequence. Left:
Posterior prediction errors as a function of the number of observations used
for inference for the three value functions: $(-3, -15, -1)$ solid, $(0, 5, 0)$ 
dashed and $(-20, 0, 1)$ dash-dot.}
\label{fig:tetris_board_configs}
\end{figure}

The aim of the first numerical experiment is to verify that it is possible to
recover
a value function and perform accurate prediction from data when the truth is
known. We consider three different value functions $%
(-3,-15, -1)$, $(0, 5, 0)$ and $(-20, 0, 1)$. These value functions were
chosen for purposes of exposition; the corresponding optimal policies lead
to qualitatively distinct styles of play. Snap-shots of typical board
configurations under play according to the action generation model for each
of these value functions are given in the top row of Figure
\ref{fig:tetris_board_configs}. 
The first value function, $(-3, -15, -1)$, led
to an \textquotedblleft efficient\textquotedblright\ style of play in which
the upper region of the board is rarely occupied. The second value function,
$(0, 5, 0)$, yields a policy which encloses many unoccupied spaces, leading
to the distinctive zig--zag pattern displayed in the second columns of
Figure \ref{fig:tetris_board_configs}. The third value function, $(-20, 0, 1)
$, 
corresponds to a policy which tends to produce
\textquotedblleft towers\textquotedblright\ of occupied squares.

For each of the three value functions, $500$ observations (state/action
pairs) were generated according to the model \eqref{eq:tetris_obs_gen} with
the state updated according to \eqref{eq:tetris_mdp_dynamics}. During
generation of the data, if the game terminated it was immediately restarted.
For the value function $(-3, -15, -1)$, termination did not occur within $%
500 $ time steps of the game. For the other two, termination typically
occurred after $10$ to $20$ time steps so the full data record of length $500
$ consisted of the concatenation of several data sets. In all three cases,
the first $100$ observations were reserved for inference and the remaining $%
400$ used for assessment of predictive performance.

\begin{figure}[h!]
\centering
\includegraphics[width=0.49\textwidth]{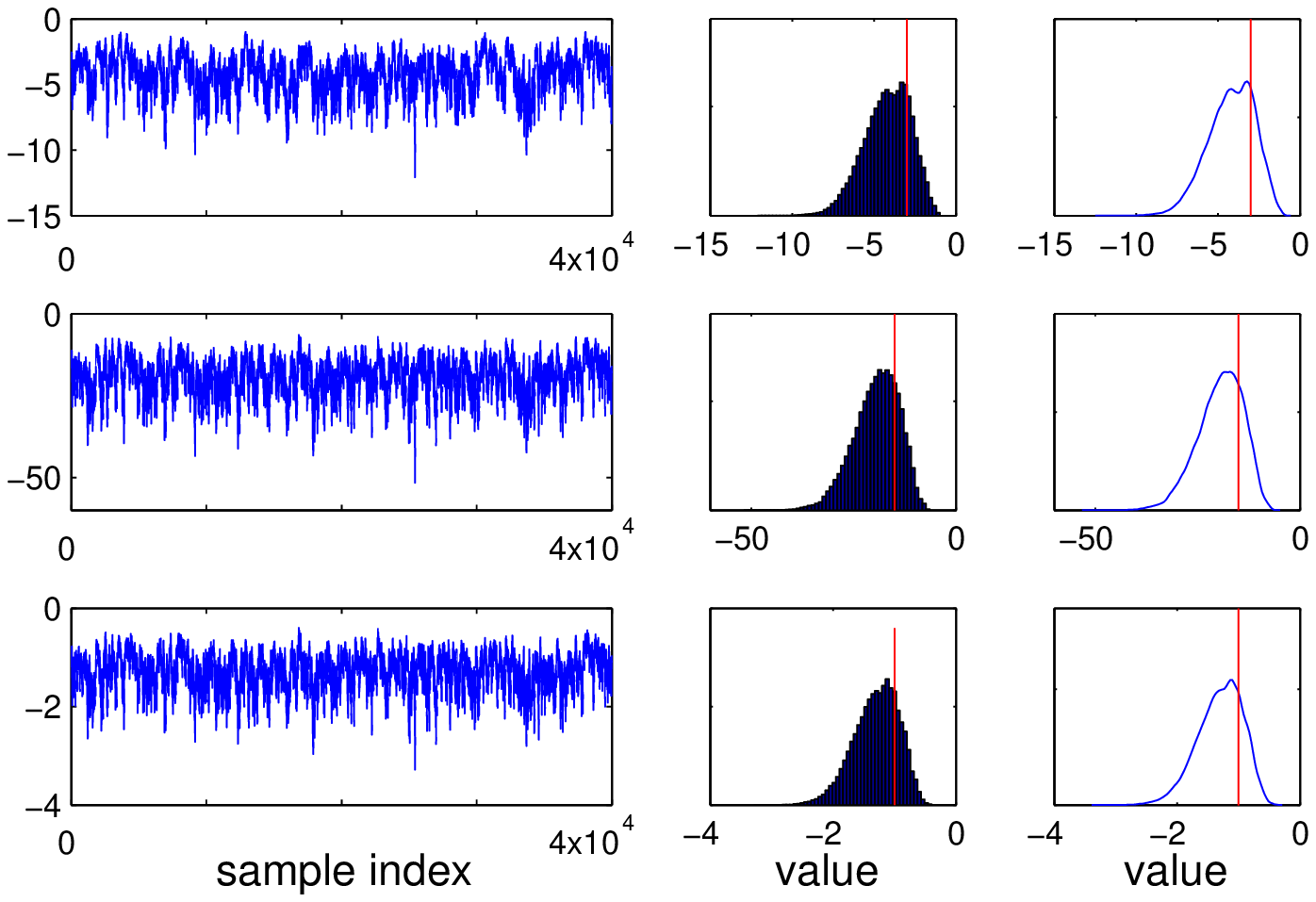} %
\includegraphics[width=0.49\textwidth]{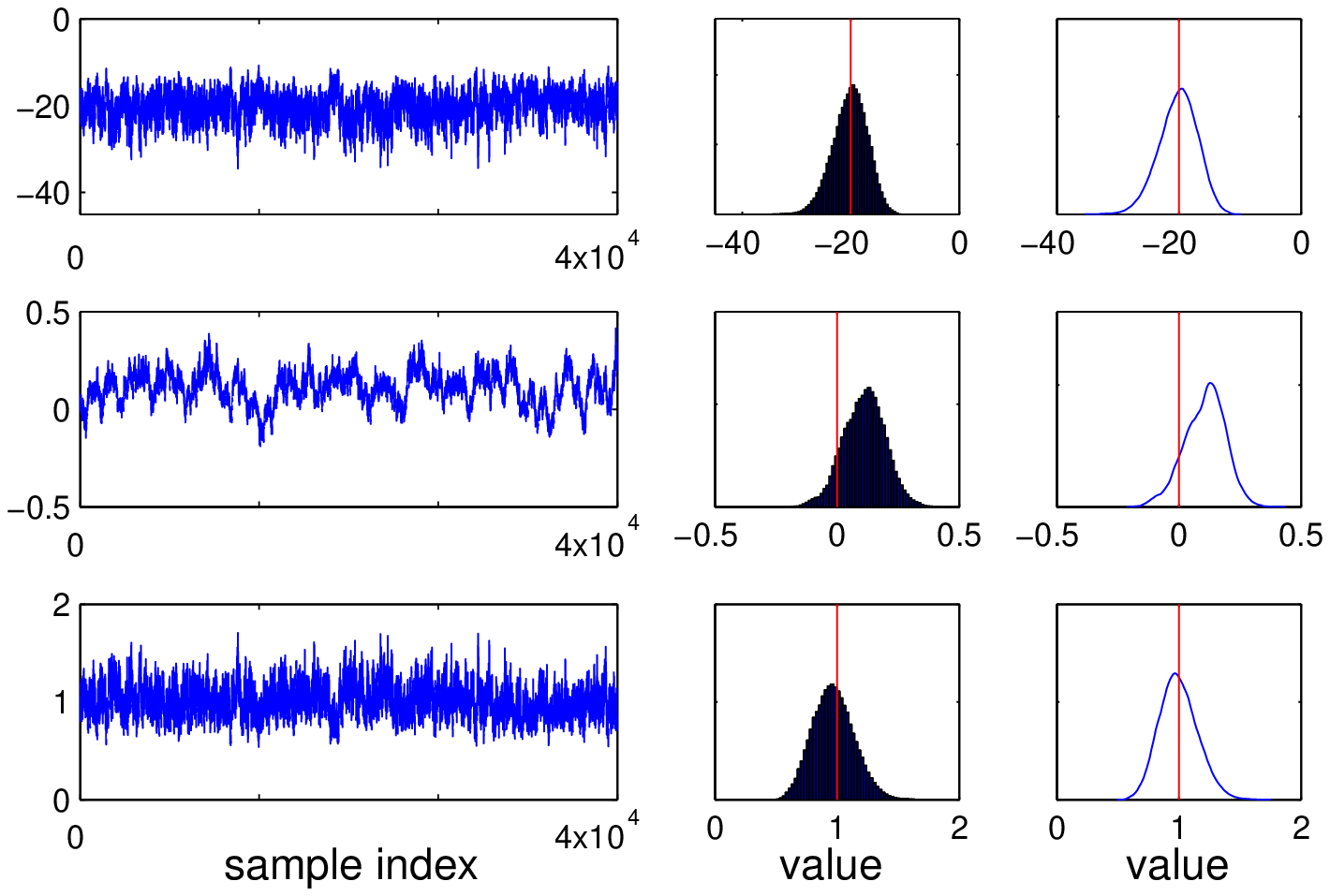}\newline
\includegraphics[width=0.49\textwidth]{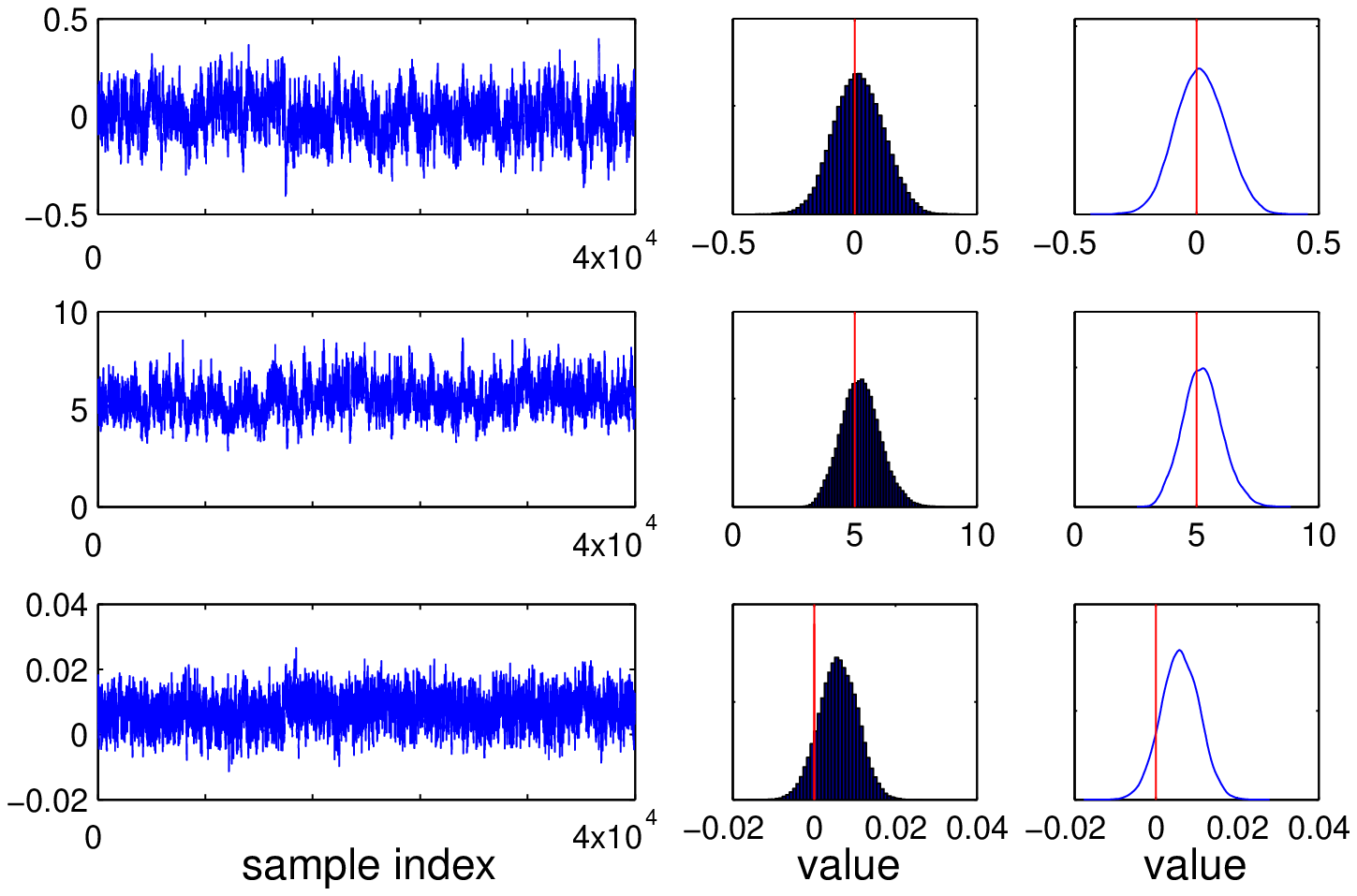} %
\includegraphics[width=0.49\textwidth]{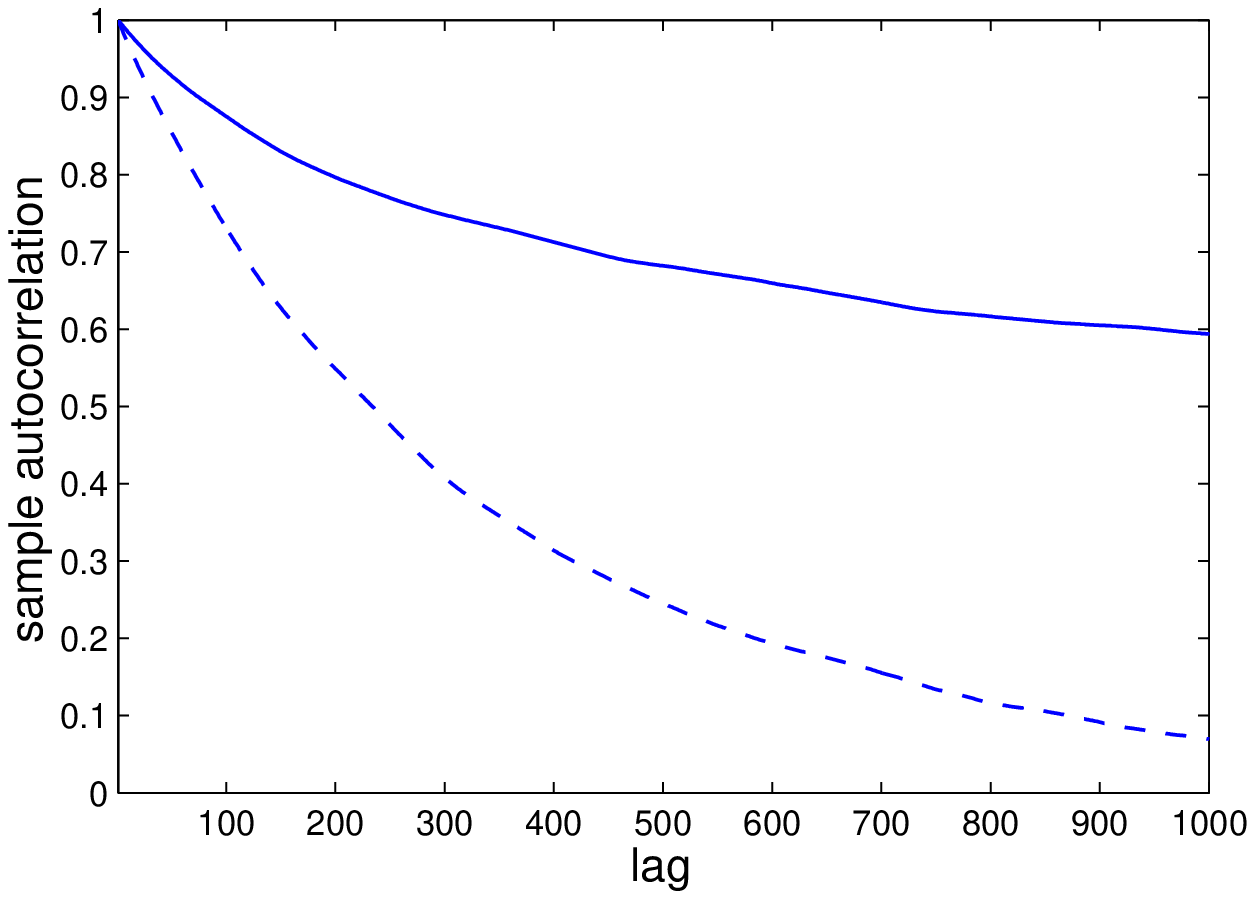}
\caption{Experiment 1. PX-DA post burn-in trace plots of $4\times 10^5$
samples, histograms and kernel density estimates of the posterior marginal
distributions corresponding to $50$ observations for the three value
functions. Top left: value function $(-3, -15, -1)$. Top right: $(-20, 0, 1)$%
. Bottom left: $(0, 5, 0)$. True values are shown with vertical lines.
Bottom right: auto-correlation as a function of lag of the first component
of $V$ for PX-DA (dashed) and DA (solid) algorithms in the case of the true
value function $(-3, -15, -1)^{T}$, from $4\times 10^5$ post burn-in
samples. }
\label{fig:tetris_exp1_traces}
\end{figure}

For each value function the PX-DA algorithm, incorporating the
Metropolis-Hastings kernel, was run independently targeting the posterior
distributions corresponding to the first $10$, $20$, $50$ and $100$
observations. In each case the algorithm was run for $5\times 10^{5}$
iterations, with a burn in of $10^{4}$ iterations. The Metropolis-Hastings
acceptance rate was found to be between $0.5$ and $0.9$ in all cases. The
parameters of the model were set to $\kappa =2500$ to give a relatively
uninformative prior over the value function, and for the prior on the
parameter $z_{1}$, $a=3$ and $b=10^{5}$. For these tuned values of $a$ and $%
b $, using an improper prior over $z_{1}$ led to negligible improvements in
performance. Post-burn in trace plots, histograms and kernel density
estimates are shown in Figure \ref{fig:tetris_exp1_traces} along with the
true value function values for the case of inference from $50$ observations.
In all cases, the posterior marginals have significant mass in the
neighborhood of the true value function values.

Figure \ref{fig:tetris_exp1_traces} also shows the autocorrelation for one
component of one of the value function, from the output of the PX-DA and
standard DA algorithms. This indicates that the PX-DA algorithm yields a
significantly lower autocorrelation than the standard DA scheme.

Figure \ref{fig:tetris_board_configs} shows the predictive performance in
terms of the prediction error $\mathcal{E}_a$ defined in equation %
\eqref{eq:error_def} as a function of the number of observations used for
inference. In all cases $\mathcal{E}_a$ was computed using the remaining $%
400 $ observations, i.e. in \eqref{eq:error_def} $T=400$. These results
verify that the predictive performance improves as the number of
observations used for inference increases.

The qualitative characteristics of play according to the three true value
functions and according to the posterior predictions are also summarized in
Figure \ref{fig:tetris_board_configs}. In this Figure, the top row shows
snap-shots of board configurations. The bottom row
shows snap-shots of play according to posterior predicted actions (with
inference based on $50$ observations) for a \emph{different} block sequence $%
\{\eta_k\}$ and with the state updated according to
\begin{align*}
\zeta_{t+1} =\psi\left(\zeta_{t},\eta_{t},\widehat{A}_{t}^{\text{MAP}%
}(\zeta_k, \eta_k)\right), & &\eta_{t+1}
\sim\mathcal{U}(1,2,...,7).
\end{align*}
These results indicate that the predicted actions result in a style of play
which is qualitatively similar to that obtained from actions generated
according to the true value function.

Lastly, with moves played according to $\hat{A}_t^{\text{MAP}}$ based on
inference from the $100$ observations generated using the value function
$(-3,-15,-1)$, termination of the game did not occur within $250$ time steps in
$80$ out of $100$ trials. In this sense, play according to
$\hat{A}_t^{\text{MAP}}$ compared quite well with play according to $A_t$, where
no termination occurred within $500$ time steps.
\subsubsection{Experiment 2}
\begin{figure}[h!]
\centering
\includegraphics[width=0.49\textwidth]{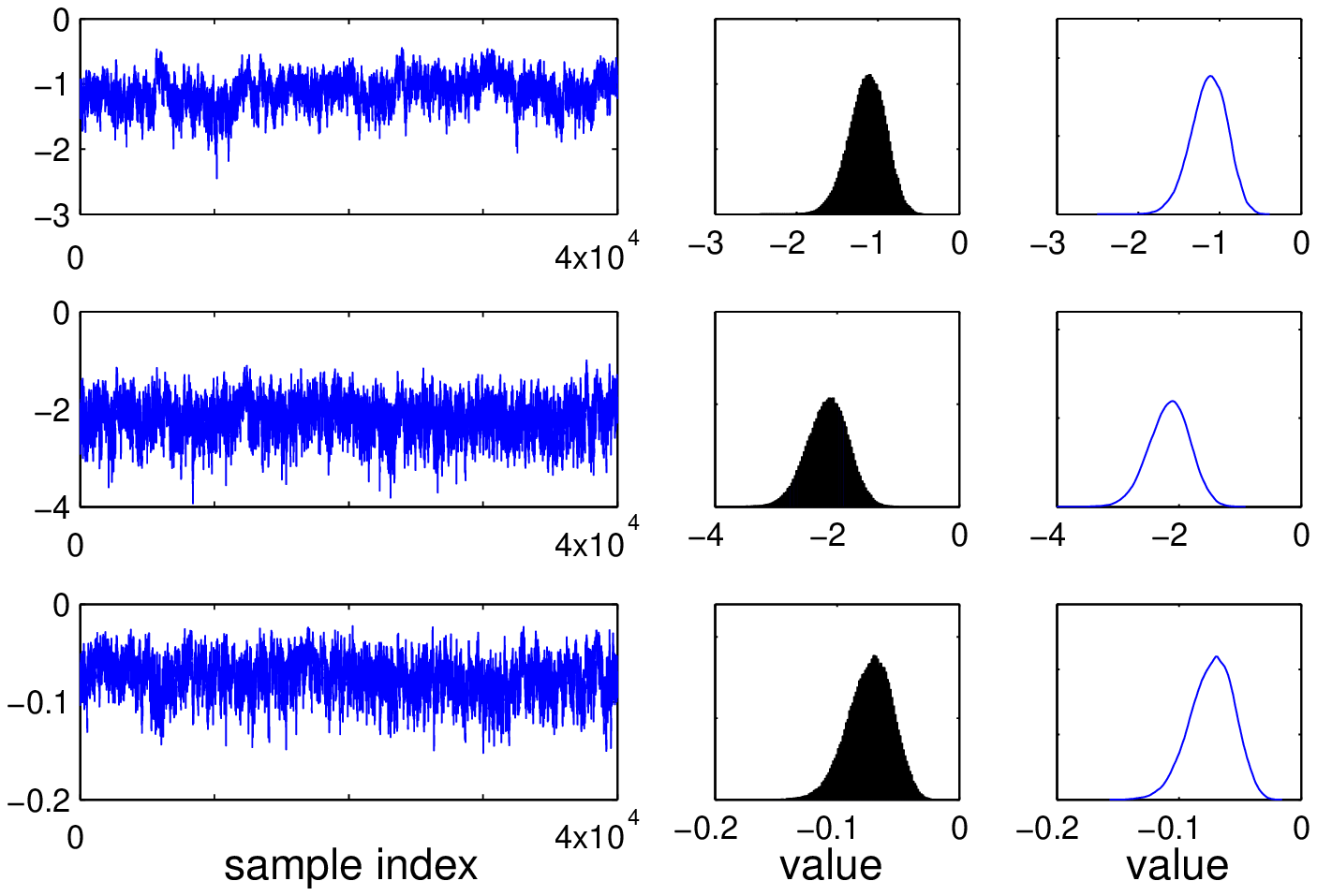} %
\includegraphics[width=0.49\textwidth]{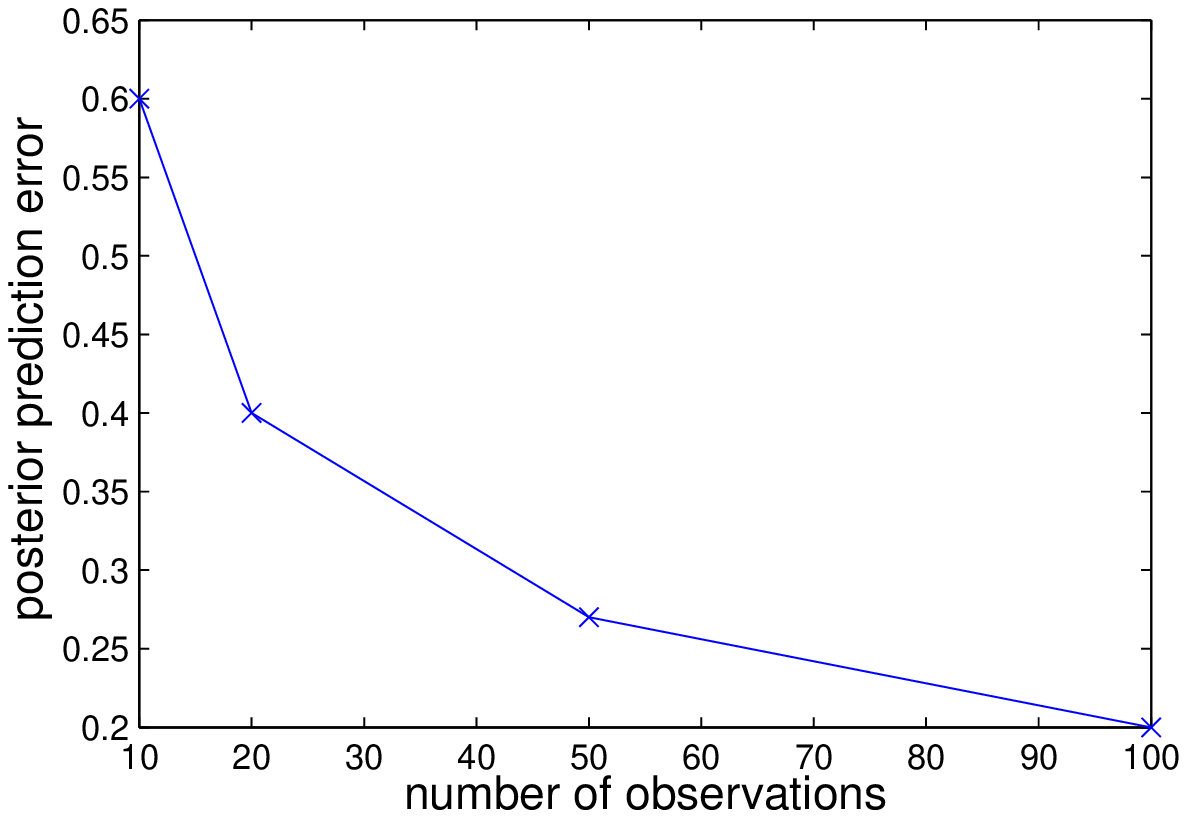}
\caption{Left: trace plots, histograms and kernel density estimates for of
posterior marginals for the three components of the value function.
Inference based on $50$ observations. Right: Posterior prediction error as a
functions of the number of observations used for inference.}
\label{fig:tetris_results_me}
\end{figure}

The aim of the second experiment is to demonstrate inference and prediction from
a data set
of a human player, i.e. in this case the true value function is
unknown. The game was played for $500$ iterations and again, the
first $100$ observations were reserved for inference and the
subsequent $400$ observations were reserved for assessment of
predictive performance. The PX-DA algorithm was run using the same
settings as in Experiment 1. Again, the Metropolis-Hastings
acceptance rate was found to be between $0.5$ and $0.9$. Trace
plots,
histograms and kernel density estimates are displayed in Figure \ref%
{fig:tetris_results_me} for the case of inference from $50$ observations.
Figure \ref{fig:tetris_results_me} also shows the empirical action error as
function of the number of observations used for inference. The result
indicates that even with three basis functions, it is possible to capture
significant information about the player's policy.

\subsubsection{Experiment 3}
In the third experiment, the play of a human was recorded under time-pressure:
at each iteration of the game a fixed time was allocated for an action to be
chosen. The aim of this experiment was to validate the statistical treatment of
the problem, by exhibiting the influence of the amount of data recorded on
inferences drawn about a player's action preferences. The experiment concerns a
situation in which the amount of data recorded is driven by the speed at which
the player is forced to play; upon the appearance of a block at the top of the
board, the player was allowed $\tau$ seconds to decide how to move the block. If
this time limit was exceeded, no action was recorded and the board was updated
by allowing the block to fall without any rotation or translation.

\begin{figure}[h!]
\centering
\includegraphics[width=0.60\textwidth]{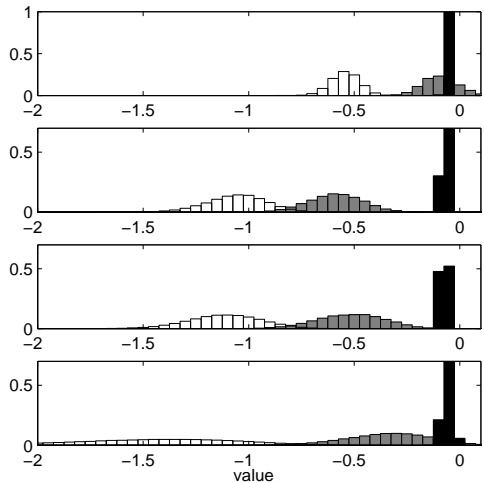} %
\caption{Experiment 3. Histogram approximations of posterior
marginals from MCMC output of $250\times10^3$ samples. Each panel
corresponds to a different constraint on the time $\tau$ allowed
per decision. Top to bottom: $\tau=10,5,3,1$ seconds. The player
was presented with $100$ blocks in total and made
respectively/Users/sumeetpalsingh/Desktop/untitled folder/refs.bib
$100,72,55,17$ decisions within the time constraints. White, grey
and black shaded histograms correspond to the three components of
the value function.} \label{fig:tetris_atc}
\end{figure}

Figure \ref{fig:tetris_atc} shows histograms of post-burn-in MCMC samples
approximating posterior marginals. Each panel corresponds to a different value
of
the time-pressure parameter $\tau$ (see caption for details). For each of the
four values of $\tau$, the player was presented with the same sequence of $100$
blocks. The hyper-parameters were set to the same values as in Experiment 2. The
results indicate that, as the time allocated for decision making was increased,
the data provide more information about their action preferences, and this is
manifested in the concentration of the posterior marginals.  A  striking feature
is the difference between the top two panels, especially in terms of mode
locations. These two panels correspond respectively to $\tau=10$ and $\tau=5$
seconds for decision making. The player made $28$ more decisions within the
allocated time in the former case than in the latter, evidently leading to
differences in posterior distributions over components of the value function.

\section{Conclusion}

Our approach to inferential computation, based on an MCMC scheme, is well suited
to the situation in which one is presented with a batch of state/action data. In
some situations, it may be that data actually arrive gradually over time, in
which case one is faced with the computational task of approximating a sequence
of posterior distributions, defined as the data become available. Sequential
Monte Carlo methods \citep{chopin2002sequential,del2006sequential} are amenable
to this kind of sequential inference computations and a possible extension of
the work presented here is to develop such methods for the class of models we
consider.
Recently, \citet{ZhS12} have applied the same probabilistic model and PX-DA
sampler developed in this paper 
to Microsoft's skill-based ranking model. The goal is to estimate the joint
probability distribution of the skills of all players (where the skill of each
player is represented by a real number) from the observation of the outcomes of
multiple games
involving subsets of these players. Preliminary results indicate that the PX-DA
sampler is more accurate in predicting the outcome of games involving closely
ranked players compared to Microsoft's variational Bayes approach (called
TrueSkill.) This research is being developed further by the authors.

\section*{Acknowledgements}

The second author is partially supported  
by the ANR grant ANR-008-BLAN-0218 “BigMC” of the French Ministry
of research. 


\section{Appendix}
\subsection{Proof of Proposition \ref{prop:scaleAndtranslate}}
The proof is essentially that of \citet[Proposition 3]{HoM08} specialized to our
specific choice for
the PX-DA move which comprises of both a scale and translation.

An operation on $\mathbb{R}_{+}\times\mathbb{R}$\ is defined as follows. For
any constants $\tilde{z}=(\tilde{z}_{1},\tilde{z}_{2}),z=(z_{1},z_{2})\in%
\mathbb{R}_{+}\times\mathbb{R}$, let
\begin{align*}
\tilde{z}z:=(\tilde{z}_{1}z_{1},\tilde{z}_{2}+\frac{z_{2}}{\tilde{z}_{1}}),&&z^{
-1}:=(z_{1}^{-1},-z_{1}z_{2}).
\end{align*}
As a consequence of these definitions,
$\varphi_{\tilde{z}}(\varphi
_{z}(y))=\varphi_{\tilde{z}z}(y)$, $\varphi_{z^{-1}}(\varphi_{z}(y))=%
\varphi_{z}(\varphi_{z^{-1}}(y))=y$ and $\varphi_{z}^{-1}(y)=\varphi
_{z^{-1}}(y)$. The following equivalences may be established by routine
integration. For any $z\in\mathbb{R}_{+}\times\mathbb{R}$ and integrable
functions $h_{1},h_{2}:\mathbb{R}^{q}\rightarrow\mathbb{R}$, $g:\mathbb{R%
}_{+}\times\mathbb{R}\rightarrow\mathbb{R}$,%
\begin{equation}
\int h_{1}(\varphi_{z}(y))J_{z}(y)dy=\int h_{1}(y)dy,
\label{eq:changeOfVariables}
\end{equation}

\begin{equation}
\int_{\mathbb{R}_{+}\times\mathbb{R}}g(z\tilde{z})dz_{1}dz_{2}=\frac{1}{%
\tilde{z}_{1}}\int_{\mathbb{R}_{+}\times\mathbb{R}}g(z)dz_{1}dz_{2},
\label{eq:operationCalculation}
\end{equation}

\begin{equation}
c(\varphi_{z}(y))=\frac{c(y)}{z_{1}J_{z}(y)},
\label{eq:normConstantCalculation}
\end{equation}

\begin{equation}
\int_{\mathbb{R}_{+}\times\mathbb{R}}g(z_{1}^{-1},-z_{1}z_{2})\frac{1}{z_{1}}%
dz_{1}dz_{2}=\int_{\mathbb{R}_{+}\times\mathbb{R}}g(z_{1},z_{2})dz_{1}dz_{2}.
\label{eq:inverseCalculation}
\end{equation}
(\ref{eq:changeOfVariables}) and (\ref{eq:operationCalculation}) follow from the
change of variable formula while (\ref%
{eq:normConstantCalculation}) follows from (\ref{eq:operationCalculation})
and the fact that
\begin{equation*}
J_{z}(\varphi_{\tilde{z}}(y))=\frac{J_{z\tilde{z}}(y)}{J_{\tilde{z}}(y)}.
\end{equation*}
Then,
\begin{align*}
& \int_{\mathbb{R}_{+}\times\mathbb{R}} \left[
\int_{\mathbb{R}^{q}}h_{1}(y)h_{2}(%
\varphi_{z}(y))\frac{f_{Y}(\varphi_{z}(y))J_{z}(y)}{c(y)}%
f_{Y}(y)dy \right] dz_{1}dz_{2} \\
& =\int_{\mathbb{R}_{+}\times\mathbb{R}}\left[ \int_{\mathbb{R}%
^{q}}h_{1}(\varphi_{z^{-1}}(\varphi_{z}(y)))h_{2}(\varphi_{z}(y))\frac {%
f_{Y}(\varphi_{z}(y))J_{z}(y)}{c(\varphi_{z^{-1}}(\varphi_{z}(y)))}%
f_{Y}(\varphi_{z^{-1}}(\varphi_{z}(y)))dy\right] dz_{1}dz_{2} \\
& =\int_{\mathbb{R}_{+}\times\mathbb{R}}\left[ \int_{\mathbb{R}%
^{q}}h_{1}(\varphi_{z^{-1}}(y))h_{2}(y)\frac{f_{Y}(y)}{c(\varphi_{z^{-1}}(y))%
}f_{Y}(\varphi_{z^{-1}}(y))dy\right] dz_{1}dz_{2} \\
& =\int_{\mathbb{R}^{q}}\left[ \int_{\mathbb{R}_{+}\times\mathbb{R}%
}h_{1}(\varphi_{z^{-1}}(y))h_{2}(y)\frac{f_{Y}(y)J_{z^{-1}}(y)}{z_{1}c(y)}%
f_{Y}(\varphi_{z^{-1}}(y))dz_{1}dz_{2}\right] dy \\
& =\int_{\mathbb{R}^{q}}\left[ \int_{\mathbb{R}_{+}\times\mathbb{R}%
}h_{1}(\varphi_{z}(y))h_{2}(y)\frac{f_{Y}(y)J_{z}(y)}{c(y)}f_{Y}(\varphi
_{z}(y))dz_{1}dz_{2}\right] dy
\end{align*}
where the final three lines are established by invoking (\ref%
{eq:changeOfVariables}), (\ref{eq:normConstantCalculation}) and (\ref%
{eq:inverseCalculation}). This establishes the stated reversibility.

\subsection{Metropolis Hastings Kernel \label{sec:MHkernel}}

For each $i=1,...,T$, one must sample $W_{i}$ from the truncated Gaussian given
in 
\eqref{eq:step1Density}. The procedure for performing this step is discussed
below for $W_{1}$, with the subscript omitted from the notation. 

Let $W(i)\sim\mathcal{N}(\mu(i),1)$ independently, $i=1,\ldots,M$. The aim
is to sample the scalar random variables $W(i)$'s conditional on the event $%
W(i)<W(l)$, for a fixed $l$ and all $i\neq l$. Without loss of generality, take
$l=1$.
The corresponding distribution for the $W(i)$'s may be decomposed as
follows. The marginal density of $W(1)$ is
\begin{equation*}
p(w(1))\propto p_{u}(w(1))=\mathcal{N}(w(1);\mu(1),1)\prod_{i=2}^{M}%
\Phi(w(1)-\mu(i)),
\end{equation*}
and, conditional on $W(1)=w(1)$, {$W(i)|\{W(1)=w(1)\}\sim\mathcal{TN}%
_{(-\infty,w(1)]}(\mu(i),1)$} for $i=2,\ldots,M$, independently, where $\Phi$
denotes the cumulative distribution function of \ $\mathcal{N}(0,1)$, and $%
\mathcal{TN}_{[a,b]}(m,s^{2})$ stands for the $\mathcal{N}(m,s^{2})$
distribution truncated to the interval $[a,b]$.

Several efficient algorithms exists for sampling from a truncated Gaussian
distribution, see \citet{chopin2011fast}. We focus on the marginal of $W(1)$. We
derive an efficient independent Metropolis-Hastings step for $W(1)$ based on a 
$\mathcal{N}(m,s^{2})$ proposal distribution. The acceptance rate reads:
\begin{equation*}
1 \wedge\frac{p_{u}(w^{\prime}(1))\mathcal{N}(w(1);m,s^{2})}{p_{u}(w(1))
\mathcal{N} (w^{\prime}(1);m,s^{2})}
\end{equation*}
where $w(1)$ and $w^{\prime}(1)$ denote, respectively, the current value and
the proposed value $W^{\prime}(1)\sim\mathcal{N}(m,s^{2})$. The main issue
is to derive a method for calculating a good Gaussian approximation
$\mathcal{N}(m,s^{2})$
 of $p(w(1))$.

The Gaussian approximation $\mathcal{N}(m,s^{2})$ is obtained iteratively. At
each
iteration, we use the following crude approximation: the function $\Phi(x)$ is 
replaced by constant one for $x>0$, and by function $\mathcal{N}(x;0,1)$ 
for $x<0$. The latter approximation is justified by the fact that, for
$x\rightarrow -\infty$, 
$x\Phi(x)/\mathcal{N}(x;0,1)\rightarrow -1$ quickly. 

At first iteration, set $(m,s^{2})=(mu(1),1)$. Then 
repeat the following steps: select the factor $i$ with
largest $\mu(i)$ and multiply the current Gaussian approximation $\mathcal{N}%
(x;m_{0},s_{0}^{2})$ by either the density $\mathcal{N}(x;\mu(i),1)$ if $%
\mu(i)>m_{0}$, or by $1$ otherwise. Discard factor $i$ and repeat this
procedure until all $M-1$ factors have been accounted for. Set $(m,s^{2})$\
to be the mean and variance of this resulting proposal.

To refine this proposal, perform several Newton-Raphson iterations for
finding the mode and the curvature of the mode of $\log p(w(1))$ by using $%
(m,s^{2})$\ as the starting values. All these operations take very little
time, and leads to an acceptance rate close to one in most cases. This
program is available upon request.

\subsection{Implementing Step 2 of Algorithm \ref{alg:inferValue}
\label{sec:algInferValStep2}}
The density (\ref{eq:step2Density}) can be written as
\begin{align}
& \mathcal{N}\left( v;\mathbf{0}_{N-1},\kappa I_{N-1}-\kappa N^{-1}\mathbf{1}%
_{N-1}\mathbf{1}_{N-1}^{\text{T}}\right) \prod \limits_{i=1}^{T}\mathcal{N}%
(w_{i}^{\prime}-R_{i}\sqrt{z_{1}}\left( v+z_{2}\mathbf{1}_{N}\right) ;%
\mathbf{0}_{M},z_{1}I_{M})  \notag \\
& \times\mathcal{N}(z_{2};0,\kappa N^{-1})IG(z_{1};a,b).
\label{eq:sampleVandZ_2}
\end{align}
By implementing the change of variable
\begin{align}
& (v(1),\ldots,v(N-1),z_{2})  \notag \\
& \rightarrow(u(1),\ldots,u(N))=\sqrt{z_{1}}\left( \left[ v(1),\ldots
,v(N-1),-\sum_{i=1}^{N-1}v(i)\right] +z_{2}\mathbf{1}_{N}^{\text{T}}\right) ,
\label{eq:changeOfVar}
\end{align}
(\ref{eq:sampleVandZ_2}) becomes
\begin{align*}
& \mathcal{N}\left( u;\mathbf{0}_{N},\kappa z_{1}I_{N}\right)
\prod
\limits_{i=1}^{T}\mathcal{N}(w_{i}^{\prime}-R_{i}u;\mathbf{0}_{M},z_{1}I_{M})
\times IG(z_{1};a,b).
\end{align*}
Sampling $(U,Z_{1})$ is now straightforward: 
\begin{equation*}
Z_{1}\sim IG(\frac{TM}{2}+a,b+SSR/2+H/2),\quad U|Z_{1}=z_{1}\sim \mathcal{N}(%
\frac{1}{z_{1}}S^{-1}\widetilde{R}^{\text{T}}\widetilde{w},S^{-1}).
\end{equation*}
where
\begin{align*}
SSR  =\widetilde{w}^{\text{T}}\widetilde{w}-\widetilde{w}^{\text{T}}%
\widetilde{R}(\widetilde{R}^{\text{T}}\widetilde{R})^{-1}\widetilde {R}^{%
\text{T}}\widetilde{w}, &&
u_{LS}  =(\widetilde{R}^{\text{T}}\widetilde{R})^{-1}\widetilde {R}^{\text{T%
}}\widetilde{w}, \\
H  =u_{LS}^{\text{T}}(I_{N}\kappa+(\widetilde{R}^{\text{T}}\widetilde {R}%
)^{-1})^{-1}u_{LS}, &&
S  =I_{N}z_{1}^{-1}\kappa^{-1}+z_{1}^{-1}(\widetilde{R}^{\text{T}}%
\widetilde{R}).
\end{align*}
and $\widetilde{w}^{\text{T}}=\left[ (w_{1}^{\prime})^{\text{T}%
},\ldots,(w_{T}^{\prime})^{\text{T}}\right] $, $\widetilde{R}^{\text{T}}=%
\left[ R_{1}^{\text{T}},\ldots,R_{T}^{\text{T}}\right] $. Here $u_{LS}$\
refers to the least squares estimate of $u$ and $SSR$ is the minimum
mean-squared error. To recover $(V,Z_{2})$ from $(U,Z_{1})$,
let $u$ denote the sampled random vector $U$, then 
$(Z_{2},V)$ is obtained as  $\left(
z_{1}^{-1/2}\frac{\boldsymbol{1}_{N}^{\text{T}%
}u}{N},z_{1}^{-1/2}(u-\frac{\boldsymbol{1}_{N}^{\text{T}}u}{N}\boldsymbol{1}%
_{N})\right) $.
\bibliographystyle{apalike}
\bibliography{refs}

\end{document}